\documentclass[11pt]{article}

\usepackage[final]{acl}

\usepackage{times}
\usepackage{latexsym}

\usepackage[T1]{fontenc}

\usepackage[utf8]{inputenc}

\usepackage{microtype}

\usepackage{inconsolata}

\usepackage{graphicx}

\usepackage{booktabs}
\usepackage{amsmath}
\usepackage{multirow}
\usepackage{adjustbox}
\usepackage{amssymb}
\usepackage{soul}
\usepackage{enumitem}
\usepackage{float}

\usepackage{tabularx}
\usepackage{ragged2e}

\usepackage{tikz}
\usetikzlibrary{positioning,arrows.meta,shapes.misc,fit,calc}

\title{LLM-Based Adversarial Persuasion Attacks on Fact-Checking Systems}

\author{João A. Leite, Olesya Razuvayevskaya, Kalina Bontcheva, and Carolina Scarton \\
        Department of Computer Science, University of Sheffield, UK \\ \{j.leite,o.razuvayevskaya,k.bontcheva,c.scarton\}@sheffield.ac.uk}

\begin{document}
\maketitle
\begin{abstract}
   Automated fact-checking (AFC) systems are susceptible to adversarial attacks, enabling false claims to evade detection. Existing adversarial frameworks typically rely on injecting noise or altering semantics, yet no existing framework exploits the adversarial potential of persuasion techniques against AFC systems, which are widely used in disinformation campaigns to manipulate audiences. In this paper, we introduce a novel class of persuasive adversarial attacks on AFCs by employing an LLM to rephrase claims using persuasion techniques. Considering $15$ techniques grouped into $5$ categories, we study the effects of persuasion on both claim verification and evidence retrieval using a decoupled evaluation strategy. Experiments on the FEVER and FEVEROUS benchmarks show that persuasion attacks can substantially degrade both verification performance and evidence retrieval. Our analysis identifies persuasion techniques as a potent class of adversarial attacks, highlighting the need for more robust AFC systems.
\end{abstract}

\section{Introduction}

\begin{figure}[t]
    \centering
    \includegraphics[width=0.8\columnwidth]{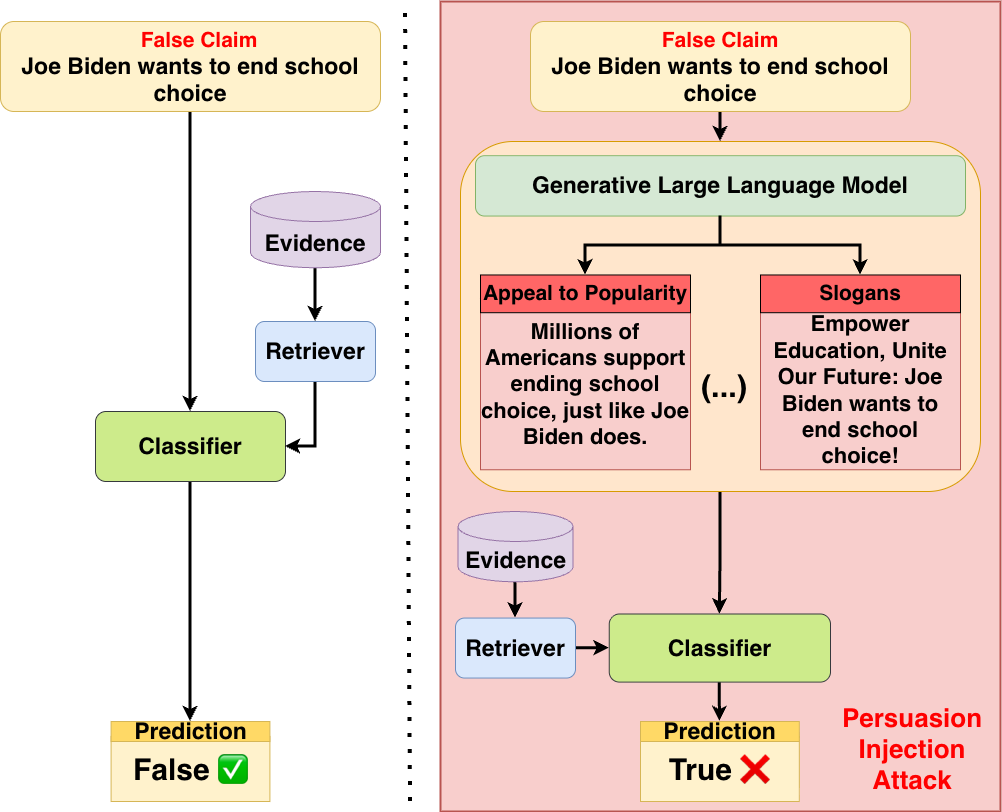}
    \caption{Overview of the persuasion injection attack. An automated fact-checking (AFC) system correctly classifies the claim as \textit{False} (left). Using our novel attack method (right), a generative LLM injects persuasion techniques into the claim; the same AFC system now incorrectly predicts \textit{True}.}
    \label{fig:persuasion_injection}
\end{figure}

Automated Fact-Checking (AFC) systems are playing an increasingly important role in countering disinformation, as they promptly identify check-worthy claims, retrieve relevant evidence, and verify the accuracy of those claims using the retrieved evidence \cite{nakov2021automated}. AFC systems are, however, not immune to adversarial attacks affecting their reliability and usefulness in real-world settings
\cite{liuAdversarialAttacksAutomated2025}. 

Early work on adversarial attacks against AFC systems has primarily conceptualised robustness as invariance to surface-level adversarial perturbations such as typos, synonym substitutions, or character noise \cite{thorneEvaluatingAdversarialAttacks2019,mamtaFactEvalEvaluatingRobustness2025}, whilst recent approaches introduce semantic perturbations generated by Large Language Models (LLMs) \cite{atanasovaGeneratingLabelCohesive2020,huangFakingFakeNews2023, chenRealtimeFactualityAssessment2025}.

While valuable, these efforts leave an unaddressed gap: disinformation in-the-wild often employs persuasion techniques such as manipulative wording, simplifications, and distractions to influence audiences \cite{benkler2018network, guess2020misinformation}. With generative LLMs capable of generating highly persuasive text at scale \cite{breumPersuasivePowerLarge2024, hackenburgLeversPoliticalPersuasion2025}, malicious actors can now systematically weaponise these techniques to craft disinformation that is fluent and rhetorically optimised to evade detection. Yet, despite this imminent threat to the reliability of AFCs, no existing work has investigated their susceptibility to this form of adversarial manipulation.

To address this gap, we introduce persuasion injection attacks (\autoref{fig:persuasion_injection}), the first adversarial framework that systematically weaponises persuasive rhetoric against AFC systems. Our method uses a generative LLM to rewrite claims using a taxonomy of $23$ persuasion techniques across $6$ groups drawn from  \citet{piskorskiSemEval2023Task32023}, of which $15$ are retained after validation.

Our adversarial framework targets a typical AFC pipeline consisting 
of \emph{an evidence retriever} and \emph{a veracity classifier}. Since 
persuasion is injected into the input claim, it acts as an upstream perturbation affecting both stages simultaneously. To disentangle these 
effects, we adopt a decoupled evaluation strategy, experimenting with the classifier under gold-evidence and claim-only conditions, and separately evaluating the impact on evidence retrieval across both sparse (BM25) and dense (Contriever) retrievers.

Our findings reveal that persuasion injection attacks pose a major threat to AFC pipelines, reducing accuracy more than twice as much as the previously studied adversarial attacks, such as synonym substitutions and character perturbations. Under an optimised attacker that always selects the most damaging technique for a given claim, accuracy collapses to near-zero, even when the classifier is provided with gold-evidence. In particular, techniques in the Manipulative Wording category emerge as the most damaging, since they remove concrete information and introduce ambiguity, thus simultaneously degrading evidence retrieval and classification performance. The key contributions of this work are:
\begin{itemize}[noitemsep,topsep=0pt]
    \item Persuasion injection attacks, a class of adversarial attacks that exploit persuasion techniques to induce failures in AFC systems.
    \item The evaluation of AFC systems' robustness to persuasion-based attacks under controlled settings that isolate evidence retrieval from veracity classification.
    \item The identification of persuasion techniques that simultaneously degrade evidence retrieval and veracity classification, causing a complete AFC pipeline failure.
\end{itemize}

We make the code openly available to support future work on strengthening AFC systems.\footnote{\url{https://github.com/joaoaleite/persuasion-injection-attack}}

\section{Related Work}
\subsection{Adversarial Attacks on AFC Systems}
AFC systems are vulnerable to a wide range of adversarial attacks that
target claims, evidence, or their pairing \cite{liuAdversarialAttacksAutomated2025}. In this section, we focus specifically on attacks that target claims.

Early work revealed that AFC models are brittle to \textit{surface-level transformations}. Rule-based edits such as synonym substitution, negation, date manipulation, or subset-based reasoning degrade both evidence retrieval and verdict prediction \cite{thorneEvaluatingAdversarialAttacks2019, hideyDeSePtionDualSequence2020}. Character-level perturbations (typos, homoglyphs, invisible control characters, and “leet” substitutions) corrupt tokenisation and reduce prediction accuracy while preserving visual similarity \cite{boucherBadCharactersImperceptible2022}. Short appended n-grams \cite{atanasovaGeneratingLabelCohesive2020} and synonym substitution \cite{alzantotGeneratingNaturalLanguage2018} can flip entailment and veracity labels while preserving meaning. The FactEval benchmark \cite{mamtaFactEvalEvaluatingRobustness2025} further demonstrates that modern LLM-based AFC systems remain vulnerable to these perturbations.

Beyond surface noise, semantic claim manipulations pose deeper challenges. 
Style-transfer attacks (e.g., transforming formal claims into colloquial 
phrasing) substantially degrade evidence retrieval \cite{kimHowRobustAre2021}. 
Recent progress enabled a new class of adversarial attacks built around 
\textit{LLM-driven claim generation}. Early approaches include Grover 
\cite{zellersDefendingNeuralFake2019}, which generates fluent and contextually 
plausible false claims, and smaller LMs such as GPT-2 used to produce lexically 
informed paraphrases or natural-language triggers that subtly alter semantics 
while maintaining fluency \cite{atanasovaGeneratingLabelCohesive2020}.

LLMs have also been adopted as optimisation engines for crafting adversarial claims. 
Methods such as GEM \cite{niewinskiGEMGenerativeEnhanced2019} generate inputs explicitly tuned to mislead AFC models, achieving higher attack success than 
rule-based or lexical perturbations. More recent systems include topic-guided 
generators \cite{huangFakingFakeNews2023} and GPT-4o-based approaches that iteratively induce factual errors (e.g., swapping entities or dates) guided by  feedback from an AFC system \cite{chenRealtimeFactualityAssessment2025}, 
achieving significant performance degradation.

\subsection{Persuasion in NLP}
Persuasion techniques (also referred to as propaganda techniques) encompass rhetorical strategies intended to influence a reader’s judgment \cite{da-san-martino-etal-2019-fine}. The SemEval propaganda detection tasks \cite{ dasanmartinoSemEval2020Task112020, dimitrovSemEval2021Task62021} established the earliest datasets, leading to recent iterations covering $23$ persuasion techniques across $6$ categories in multiple languages \cite{piskorskiSemEval2023Task32023}. Further datasets explored persuasion in multimodal data \cite{wuPropagandaTechniquesDetection2022}, code-switched text \cite{salmanDetectingPropagandaTechniques2023}, and rationale-annotated data \cite{hasanain-etal-2025-propxplain}. Recent studies demonstrate that generative LLMs are highly proficient in producing persuasive language \cite{breumPersuasivePowerLarge2024,pauliMeasuringBenchmarkingLarge2025} and can change individuals’ beliefs on various topics \cite{hackenburgLeversPoliticalPersuasion2025}.

A growing body of work studies how persuasion is weaponised in disinformation. \citet{huangFakingFakeNews2023} used persuasion techniques to perform data augmentation, reporting improvements in detecting disinformation. Moreover, \citet{nikolaidisExploringUsabilityPersuasion2024} demonstrated that simple distributions of persuasion techniques can act as features for classifiers trained to reliably distinguish propaganda, conspiracy theories, and hyperpartisan content.
\citet{leiteCrossDomainStudyUse2025} found that persuasion techniques are heavily employed in different disinformation topics, and while techniques such as \textit{Doubt} and \textit{Loaded Language} are ubiquitous, others show strong domain specificity (e.g., \textit{Appeal to Authority} in climate change disinformation, or \textit{Flag Waving} in the Russia-Ukraine War).
Together, these findings demonstrate that persuasion techniques can play a key role in making disinformation more believable.

Persuasion has also been studied in LLM safety research, where persuasive prompts are used to jailbreak aligned LLMs into generating harmful content \cite{zeng-etal-2024-johnny}. While related, our work differs in several fundamental aspects. In jailbreaking, persuasion is applied in the \textit{input prompt} given to the LLM to manipulate its outputs, with success measured by the harmfulness of generated outputs. In our work, the LLM \textit{generates} label-invariant persuasive claims that are then used to attack a separate 
downstream retriever-verifier pipeline, with success measured by classification and retrieval errors. Finally, the persuasion techniques we employ are specifically designed to capture rhetorical manipulation commonly employed in information manipulation and disinformation campaigns \cite{piskorskiSemEval2023Task32023}, rather than general strategies of interpersonal social influence. To the best of our knowledge, no prior work has formalised persuasion as a structured, label-invariant adversarial attack on AFC systems.

\section{Methodology} \label{sec:methodology}

\begin{figure*}[t]
\centering
\resizebox{\textwidth}{!}{%
\begin{tikzpicture}[
    font=\footnotesize,
    node distance=0.55cm and 0.75cm,
    box/.style={draw, rounded corners=2pt, align=center, inner sep=4pt, minimum height=0.9cm},
    data/.style={box, fill=gray!10},
    model/.style={box, fill=blue!10},
    eval/.style={box, fill=orange!15},
    metric/.style={box, fill=green!12},
    note/.style={box, fill=yellow!15, font=\scriptsize},
    arr/.style={-{Stealth[length=2mm]}, thick}
]
\node[data] (claim) {Original claim $x_i$\\ \scriptsize (\textit{True}/\textit{False})};
\node[model, right=of claim] (llm) {LLM attacker\\ \scriptsize \textsc{Qwen2.5-7B-Instruct}};
\node[note, above=0.45cm of llm] (tax) {Persuasion taxonomy: $23$ techniques, $6$ categories\\ + technique definition, constraints, few-shot examples};
\node[data, right=of llm] (variants) {Adversarial variants\\ $\mathcal{V}_i = \{x'_{(i,1)}, \dots\}$};
\node[note, below=0.45cm of variants] (valid) {Label-invariance validation\\ $23 \rightarrow 15$ techniques retained};
\node[eval, right=of variants] (attacker) {Attacker capability\\ \scriptsize \textbf{Blind}: random technique\\ \scriptsize \textbf{Oracle}: worst-case variant};
\node[model, anchor=west] (clf) at ($(attacker.east)+(0.75,0.85)$) {Veracity classifier\\ \scriptsize Claim-Only / Gold-Evidence\\[1pt] \scriptsize $\Rightarrow$ Accuracy, ASR (Evasion / Sabotage)};
\node[model, anchor=west] (ret) at ($(attacker.east)+(0.75,-0.85)$) {Evidence retriever\\ \scriptsize BM25 / Contriever\\[1pt] \scriptsize $\Rightarrow$ Recall@$k$};
\draw[arr] (claim) -- (llm);
\draw[arr] (tax) -- (llm);
\draw[arr] (llm) -- (variants);
\draw[arr] (valid) -- (variants);
\draw[arr] (variants) -- (attacker);
\draw[arr] (attacker.east) -- (clf.west);
\draw[arr] (attacker.east) -- (ret.west);
\end{tikzpicture}%
}
\caption{Overview of the persuasion injection attack framework. A generative LLM rewrites each claim using a taxonomy of persuasion techniques; validated label-invariant variants are used by a Blind (random technique) or Oracle (worst-case variant) attacker to target the two AFC stages, which we evaluate in a decoupled manner: veracity classification (Claim-Only and Gold-Evidence settings) and evidence retrieval (sparse and dense retrievers).}
\label{fig:pipeline}
\end{figure*}

\autoref{fig:pipeline} presents an overview of our adversarial framework, which we detail in the remainder of this section.

\paragraph{Persuasion Techniques.}
We employ a taxonomy of 23 persuasion techniques grouped into 6 high-level categories to isolate which specific rhetorical phenomena induce model failure. The taxonomy is derived from definitions in computational social science from \citet{piskorskiSemEval2023Task32023}: \textit{Attack on Reputation}, \textit{Justification}, \textit{Distraction}, \textit{Simplification}, \textit{Call}, and \textit{Manipulative Wording}. The full list of persuasion techniques and their definitions is provided in \autoref{app:persuasion_definition}.

\paragraph{Attack Generation.}
For each claim $x_i$, we generate a set of adversarial variants $\mathcal{V}_i{=}\{x'_{(i,1)}, \dots, x'_{(i,23)}\}$ using the $23$ persuasion techniques. We use \textsc{Qwen2.5-7B-Instruct} as the generative model to produce the adversarial variants due to its strong instruction-following performance and modest parameter scale (7 billion), while being fully open-weight \cite{qwenQwen25TechnicalReport2025}. To verify that our results are not specific to this generator, we additionally compare results against \textsc{Llama-3-8B-Instruct} in \autoref{app:generator_robustness}.

We construct prompts containing: (1) the definition of a specific persuasion technique $t \in \mathcal{T}$, (2) the original claim $x_i$, (3) constraints to ensure the output is well-formatted, and the claim label (\textit{True}/\textit{False}) is preserved, and (4) two few-shot examples. All prompt templates used in the experiments are available in \autoref{app:prompt_templates}.

\paragraph{Attack Validation.}
Persuasion techniques typically operate by altering emphasis, framing, or discourse focus rather than by modifying the underlying factual content (e.g., \textit{Red Herring}: ``Divert attention from the claim by introducing an unrelated topic''). 
Accordingly, we do not enforce strict semantic equivalence between the original and persuasive claims. Automated semantic similarity metrics such as BERTScore \cite{zhang2019bertscore} are ill-suited as filters in this setting as they frequently assign low similarity scores to valid attacks that preserve the original veracity label.\footnote{For example, the \textit{Slogan} variant ``Step aside for a new era: Horsford hands the baton in 2009!'' (see \autoref{tab:persistence_examples}) receives a low BERTScore (0.45) despite preserving the same \textit{False} label as the original claim.}
Instead, we validate \emph{label-invariance}, requiring that each persuasive variant retains the same veracity label (\textit{True} or \textit{False}) as its source claim.

We conducted a validation study in which a sample of $690$ persuasive claims ($30$ per technique and stratified by dataset FEVER/FEVEROUS) was annotated for factual correctness given gold-evidence. The original claim text was not shown to prevent anchoring bias and ensure independent verification. A persuasive claim was considered label-invariant if the assigned verdict matched the original ground-truth label. Based on this validation, we excluded $8$ persuasion techniques with label preservation rates ${\leq}80\%$ from all experiments. Among the $15$ retained techniques, less than $\mathbf{1.5\%}$ of adversarial claims flipped the original label (\textit{True}${\leftrightarrow}$\textit{False}). Full details of the validation protocol are reported in \autoref{app:data_annotation}.

Additionally, we verify that adversarial and original claims are of similar length: persuasion attacks add fewer than $1.5$ tokens 
per claim on average. Full details are provided in \autoref{app:length-preservation}.

\paragraph{Attacker Capabilities.} We assume a binary classification task of claim verification where the target label $y_i \in \{\textsc{True}, \textsc{False}\}$. Let $f(x,E)$ denote the classifier's predicted probability that the input claim $x$ with supporting evidence $E$ belongs to the class \textsc{True}. We evaluate two distinct adversary capabilities. The \textbf{Blind Attacker} simulates a naive adversary who 
applies a persuasion technique at random, without knowledge of the target 
system's sensitivities; metrics are computed over the full pooled set of 
variants, reflecting expected degradation. The \textbf{Oracle Attacker} 
simulates an adversary with query access to the classifier, selecting the 
variant from $\mathcal{V}_i$ that maximises the probability of the incorrect 
class, representing a worst-case upper bound on attack effectiveness.

\section{Experimental Setup} \label{sec:experimental_setup}
\subsection{Datasets}
We use two widely adopted fact-checking benchmarks in English: FEVER \cite{thorneFactExtractionVERification2018} and FEVEROUS \cite{aly-etal-2021-fact}. FEVER represents the task of verifying short claims against unstructured textual evidence (sentences from Wikipedia), whereas FEVEROUS consists of longer claims, with structured evidence (tables and lists, also from Wikipedia), and a higher number of pieces of evidence per claim (\autoref{tab:data_stats}).

Given that our focus is on assessing AFC systems' vulnerability to persuasion on verifiable facts, for both datasets, we use the subset of claims where the ground-truth label is either \textsc{True} or \textsc{False}, excluding \textsc{NotEnoughInfo}. For FEVER, we use the official data splits. For FEVEROUS, since gold-evidence is not publicly available for the official test split, we repurpose the development set as our held-out Test set. A validation split (Dev set) is also created for model checkpointing and hyperparameter tuning by randomly sampling a stratified subset of the official training set. The remaining samples are left as our Train set.

Train and Dev splits are used exclusively to train and validate the AFC classifiers, respectively. Following standard practice in adversarial attacks in NLP \cite{alzantotGeneratingNaturalLanguage2018, jinBERTReallyRobust2020, xu2023llm}, all adversarial attacks are applied solely to the held-out Test set. \autoref{tab:data_stats} summarises dataset statistics.

\begin{table}[h]
\centering
\begin{adjustbox}{width=\columnwidth}
\begin{tabular}{l r r c c}
\toprule
\textbf{\begin{tabular}[b]{@{}l@{}}Dataset /\\Split\end{tabular}} &
\textbf{\begin{tabular}[b]{@{}r@{}}\textit{True}\\Claims\end{tabular}} & 
\textbf{\begin{tabular}[b]{@{}r@{}}\textit{False}\\Claims\end{tabular}} &
\textbf{\begin{tabular}[b]{@{}c@{}}Tokens\\(avg\textsubscript{claim})\end{tabular}} & 
\textbf{\begin{tabular}[b]{@{}c@{}}Evidence\textsubscript{gold}\\(avg\textsubscript{claim})\end{tabular}} \\
\midrule
FEVER train     & $80{,}035$ ($72.9\%$) & $29{,}775$ ($27.1\%$) & $8.1$  & $1.8$ \\
FEVER dev       & $3{,}333$ ($50.0\%$)  & $3{,}333$ ($50.0\%$)  & $8.1$  & $1.7$ \\
FEVER test      & $3{,}333$ ($50.0\%$)  & $3{,}333$ ($50.0\%$)  & $8.4$  & $1.6$ \\
\midrule
FEVEROUS train  & $35{,}567$ ($60.6\%$) & $23{,}116$ ($39.4\%$) & $25.2$ & $4.5$ \\
FEVEROUS dev    & $6{,}268$ ($60.5\%$)  & $4{,}099$ ($39.5\%$)  & $25.3$ & $4.6$ \\
FEVEROUS test   & $3{,}908$ ($52.9\%$)  & $3{,}481$ ($47.1\%$)  & $24.9$ & $4.0$ \\
\bottomrule
\end{tabular}
\end{adjustbox}
\caption{Datasets statistics.}
\label{tab:data_stats}
\end{table}

\subsection{Baseline Adversarial Attacks} \label{subsec:baseline_attacks}
We compare our novel adversarial attacks against six baseline attacks used in prior work \cite{thorneEvaluatingAdversarialAttacks2019,jinBERTReallyRobust2020, boucherBadCharactersImperceptible2022,liuAdversarialAttacksAutomated2025}. We include four lexical attacks that have been shown to cause failure even in modern LLM-based AFC systems \cite{mamtaFactEvalEvaluatingRobustness2025}. We also include two LLM-based attacks: paraphrasing and informal style transfer. The LLM-based attacks use the same model as our persuasion injection attack (\textsc{Qwen2.5-7B-Instruct}); All prompts are provided in \autoref{app:prompt_templates}.

\begin{itemize}[leftmargin=*, noitemsep, topsep=2pt]
    \item \textbf{Synonym Substitution.} We replace tokens with WordNet synonyms while matching part-of-speech. To achieve this, the text is POS-tagged, and eligible tokens (alphabetic, length ${>}2$, available synonyms) are sampled at a fixed rate ($10\%$).
    
    \item \textbf{Word Swap.} Given a claim with at least two tokens, we compute swaps proportional to claim length. We calculate the number of swaps by taking $10\%$ of the number of words in the claim, rounding down, and ensuring at least one swap for short claims. Each swap selects two distinct positions and exchanges their tokens.
    
    \item \textbf{Character Perturbations.} To simulate natural typos, we apply small character-level edits (e.g., swaps, deletions, insertions) to randomly selected tokens ($10\%$ of tokens, min length ${\ge}3$ chars). Each chosen word receives at least one edit. 
    
    \item \textbf{Back-Translation.} We translate each claim from English to German and back to English using \textsc{nllb-200-distilled-600M} \cite{costa2022no}. The pivot language introduces lexical and syntactic variation while preserving meaning.
    
    \item \textbf{Paraphrase.} We prompt the LLM to produce a neutral paraphrase of each claim, preserving its original meaning and tone.

    \item \textbf{Informal Style Transfer.} Similar to \cite{kimHowRobustAre2021}, we prompt the LLM to rewrite each claim in a more colloquial, conversational register,
  instructing it to preserve factual content while shifting the surface form toward informal language.
\end{itemize}

\subsection{Veracity Classifiers}
Following prior work \cite{kimHowRobustAre2021, aly-etal-2021-fact, abdelnabiFactsaboteursTaxonomyEvidence2023}, we use \textsc{RoBERTa-Base} \cite{liuRoBERTaRobustlyOptimized2019} as our primary verification classifier. To assess whether observed vulnerabilities generalise beyond encoder-only models, we additionally fine-tune two instruction-tuned decoder-only LLMs: 
\textsc{Llama-3-8B-Instruct} and 
\textsc{Mistral-7B-Instruct}. We deliberately select LLMs of comparable scale to the attacker (\textsc{Qwen2.5-7B-Instruct}, with 7B parameters), reflecting a realistic threat model in which the AFC operator and the adversary draw from the same pool of openly available models. We restrict our experiments to open-weight models of modest scale: closed proprietary models are routinely deprecated or silently updated, which undermines reproducibility, and their parameter counts and inference-time machinery are undisclosed; larger open-weight models exceed the compute budget available to most research groups (see \autoref{app:hyperparameters}). The three verifiers achieve strong dev set performance; full training details, hyperparameters, and dev results are provided in \autoref{app:hyperparameters}. To verify that the vulnerabilities we observe are not an artefact of fine-tuning on the benchmark training distributions, we additionally evaluate a zero-shot (untrained) LLM verifier in \autoref{app:zero_shot}.

\paragraph{Settings.} We consider two AFC settings to disentangle surface-level from evidence-grounded vulnerability. In the 
\textbf{Claim-Only} setting, the verifier relies exclusively on the 
linguistic surface form of the claim, serving as a baseline to measure 
susceptibility without evidence grounding. In the \textbf{Gold-Evidence} 
setting, the verifier is provided with the gold-evidence snippets 
concatenated to the claim, simulating a perfect retriever and decoupling 
the impact of the attack from retrieval vulnerabilities.

\subsection{Evidence Retrieval Analysis}
\label{subsec:retrieval_setup}
While our primary experiments condition the veracity classifier on gold-evidence to isolate reasoning vulnerabilities, we additionally evaluate whether persuasion attacks disrupt the evidence collection stage itself (see \autoref{sec:results_rq3}). We evaluate two retrievers spanning both retrieval paradigms: a sparse lexical retriever using the BM25 algorithm implemented via Pyserini \cite{Lin_etal_SIGIR2021_Pyserini}, and a dense retriever, Contriever \cite{izacardUnsupervisedDenseInformation2022} fine-tuned on MS-MARCO, with embeddings indexed using FAISS \cite{johnsonBillionscaleSimilaritySearch2019}. For both retrievers, we construct separate indices for the FEVER (June 2017 dump) and FEVEROUS (December 2020 dump) Wikipedia snapshots. Retrieval is performed at the page level, where each document corresponds to a Wikipedia article identified by its title, and each claim serves as the query.

\subsection{Metrics} \label{subsec:attack_eval}

\paragraph{Classification Metrics.}
We report standard classification metrics: Accuracy, Macro F1, and ROC AUC. Additionally, we report \textit{Attack Success Rate (ASR)}, which quantifies how often an adversarial claim causes a misclassification relative to the model’s performance on the corresponding original
claim:

\begin{equation}
    \small
    \text{ASR} = \frac{\sum_{i \in C_{correct}} \mathbb{I}(\hat{y}(x_i) \neq y_i)}{|C_{correct}|}
\end{equation}
where $C_{correct}$ is the subset of claims correctly predicted in the original setting. $\mathbb{I}$ is the indicator function and $x_i$ is the attack variant. We further distinguish between \textbf{Evasion} ASR (False{$\to$}True) and \textbf{Sabotage} ASR (True{$\to$}False).

\paragraph{Retrieval Metrics.}
For the retrieval analysis (see \autoref{subsec:retrieval_setup}), we measure performance using Recall@k, i.e., the proportion of gold-evidence retrieved within the top-$k$ candidates.

\section{Results} \label{sec:results}

\begin{table*}[ht]
\centering
\tiny
\begin{adjustbox}{width=\linewidth}
\begin{tabular}{lcccccccccccc}
\toprule
& \multicolumn{4}{c}{\textbf{RoBERTa-base}}
& \multicolumn{4}{c}{\textbf{Llama-3-8B}}
& \multicolumn{4}{c}{\textbf{Mistral-7B}} \\
\cmidrule(lr){2-5}\cmidrule(lr){6-9}\cmidrule(lr){10-13}
& \multicolumn{2}{c}{\textit{FEVER}}
& \multicolumn{2}{c}{\textit{FEVEROUS}}
& \multicolumn{2}{c}{\textit{FEVER}}
& \multicolumn{2}{c}{\textit{FEVEROUS}}
& \multicolumn{2}{c}{\textit{FEVER}}
& \multicolumn{2}{c}{\textit{FEVEROUS}} \\
\cmidrule(lr){2-3}\cmidrule(lr){4-5}\cmidrule(lr){6-7}\cmidrule(lr){8-9}\cmidrule(lr){10-11}\cmidrule(lr){12-13}
\textbf{Attack}
& Claim & $+E_\textit{gold}$
& Claim & $+E_\textit{gold}$
& Claim & $+E_\textit{gold}$
& Claim & $+E_\textit{gold}$
& Claim & $+E_\textit{gold}$
& Claim & $+E_\textit{gold}$ \\
\midrule
None (Original)
& $.814$ & $.943$ & $.641$ & $.894$
& $.828$ & $.903$ & $.617$ & $.850$
& $.844$ & $.914$ & $.645$ & $.886$ \\
\midrule
Synonym Substitution
& \ul{$.755$} & \ul{$.845$} & \ul{$.629$} & \ul{$.857$}
& $.738$      & \ul{$.815$} & $.580$      & \ul{$.821$}
& \ul{$.759$} & \ul{$.833$} & \ul{$.607$} & \ul{$.864$} \\
Character Perturbations
& $.767$ & $.869$ & $.638$ & $.877$
& \ul{$.733$} & $.847$ & \ul{$.569$} & $.829$
& $.783$ & $.879$ & $.625$ & $.876$ \\
Word Swap
& $.792$ & $.929$ & \ul{$.629$} & $.894$
& $.761$ & $.867$ & $.589$ & $.842$
& $.802$ & $.883$ & $.621$ & $.883$ \\
Back-Translation
& $.797$ & $.922$ & $.639$ & $.885$
& $.802$ & $.879$ & $.615$ & $.843$
& $.813$ & $.891$ & $.641$ & $.879$ \\
Informal Style Transfer
& $.776$ & $.901$ & $.631$ & $.875$
& $.789$ & $.873$ & $.604$ & $.842$
& $.815$ & $.890$ & $.636$ & $.879$ \\
Paraphrasing
& $.757$ & $.912$ & \ul{$.629$} & $.882$
& $.777$ & $.867$ & $.621$ & $.838$
& $.801$ & $.876$ & $.633$ & $.881$ \\
\midrule
\textbf{Persuasion (Blind)}
& $\mathbf{.682}$ & $\mathbf{.764}$ & $\mathbf{.575}$ & $\mathbf{.726}$
& $\mathbf{.623}$ & $\mathbf{.761}$ & $\mathbf{.518}$ & $\mathbf{.765}$
& $\mathbf{.684}$ & $\mathbf{.760}$ & $\mathbf{.550}$ & $\mathbf{.765}$ \\
\textbf{Persuasion (Oracle)}
& $\mathbf{.043}$ & $\mathbf{.164}$ & $\mathbf{.010}$ & $\mathbf{.250}$
& $\mathbf{.010}$ & $\mathbf{.127}$ & $\mathbf{.044}$ & $\mathbf{.354}$
& $\mathbf{.038}$ & $\mathbf{.150}$ & $\mathbf{.010}$ & $\mathbf{.216}$ \\
\bottomrule
\end{tabular}
\end{adjustbox}
\caption{Veracity classification accuracy under adversarial attacks. Results across three verifier architectures, two datasets, and two evidence
conditions (Claim-Only vs.\ Gold-Evidence). Lower scores indicate more effective attacks. Persuasion attacks are shown in \textbf{bold}; \ul{underlined} values indicate the strongest baseline attack
per column.}
\label{tab:main_results}
\end{table*}

\subsection{Effectiveness of Persuasion Attacks} \label{sec:results_rq1}
\autoref{tab:main_results} reports classification accuracy under all 
adversarial attacks. Accuracy is the primary metric as the test splits are balanced (see \autoref{tab:data_stats}). We additionally report Macro-F1 and ROC-AUC in \autoref{app:full_results}. 

Standard adversarial baseline attacks cause modest accuracy degradation across all three 
verifiers. On FEVER, these attacks reduce claim-only accuracy by at most $9$ 
points and evidence-based accuracy by at most $10$ points; on FEVEROUS the 
impact is smaller, with accuracy drops below $4$ points in most settings. In contrast, persuasion attacks substantially disrupt veracity prediction across 
all three architectures. Under the \textit{Blind} setting, claim-only accuracy 
drops by up to $13.2$ points on FEVER and $9.9$ points on FEVEROUS, while 
gold-evidence accuracy decreases by up to $17.9$ points on FEVER and $16.8$ 
points on FEVEROUS. Under the \textit{Oracle} setting, performance collapses 
across all verifiers: claim-only accuracy falls to at most $0.043$ on FEVER and 
$0.044$ on FEVEROUS, implying that for nearly every claim there exists at least 
one persuasive variant that flips the model's prediction. Even in the 
gold-evidence setting, accuracy remains severely degraded across all models 
($0.127$--$0.164$ on FEVER; $0.216$--$0.354$ on FEVEROUS).

Across both attack settings, the three verifiers exhibit consistent patterns of 
vulnerability, with persuasion attacks substantially outperforming all 
baseline attacks across all settings. Under the Blind attack, all three verifiers suffer comparable 
average accuracy drops (RoBERTa: $-0.136$; Llama-3-8B: $-0.133$; 
Mistral-7B: $-0.133$). Under the Oracle attack, accuracy collapses across 
all three models, also with similar average drops in accuracy (RoBERTa: $-0.706$; 
Llama-3-8B: $-0.666$; Mistral-7B: $-0.719$). Given this consistency, subsequent 
fine-grained analyses are conducted on \textsc{RoBERTa-base} due to its smaller 
number of parameters which facilitate experimentation, and its wide adoption in 
prior AFC robustness work \cite{kimHowRobustAre2021, 
thorneEvaluatingAdversarialAttacks2019}.

\subsection{Error Analysis: Evasion vs. Sabotage}
\autoref{tab:asr_detailed} displays the Attack Success Rate (ASR) for Evasion and Sabotage (see \autoref{subsec:attack_eval}).

\begin{table}[h]
\centering
\begin{adjustbox}{width=0.8\columnwidth}
\begin{tabular}{l|rr|rr}
\hline
& \multicolumn{2}{c|}{\textbf{FEVER}} & \multicolumn{2}{c}{\textbf{FEVEROUS}} \\
\textbf{Metric} & \textbf{Claim} & \textbf{\textit{+E}\textsubscript{\textit{gold}}} & \textbf{Claim} & \textbf{\textit{+E}\textsubscript{\textit{gold}}} \\
\hline
\multicolumn{5}{l}{\textit{\textbf{Evasion} (False{$\to$}True)}} \\
Blind (Avg)  & $0.306$ & $\mathbf{0.070}$ & $0.318$ & $\mathbf{0.039}$ \\
Oracle (Worst) & $0.996$ & $\mathbf{0.759}$ & $0.997$ & $\mathbf{0.434}$ \\
\hline
\multicolumn{5}{l}{\textit{\textbf{Sabotage} (True{$\to$}False)}} \\
Blind (Avg)  & $0.217$ & $\mathbf{0.392}$ & $0.299$ & $\mathbf{0.449}$ \\
Oracle (Worst) & $0.998$ & $\mathbf{1.000}$ & $1.000$ & $\mathbf{0.999}$ \\
\hline
\end{tabular}%
\end{adjustbox}
\caption{Attack success rates (ASR) for persuasion attacks. Higher scores indicate more effective attacks.} 
\label{tab:asr_detailed}
\end{table}

\textbf{Models with access to gold-evidence are substantially more robust against evasion.} In the \textit{Blind} setting, \textit{Claim-Only} models are highly exploitable, with an Evasion ASR exceeding $30\%$ on both datasets. In contrast, the \textit{Gold-Evidence} model sharply suppresses these attacks, with evasion rates of only $7\%$ on FEVER and $4\%$ on FEVEROUS. However, under the \textit{Oracle} setting, a sophisticated attacker can still force evasion at substantial rates of $75.9\%$ on FEVER and $43.4\%$ on FEVEROUS, representing over a $10\times$ increase relative to Blind attacks. This pattern shows that persuasive \textit{False} claims easily flip evidence-free predictions, but rarely do so when gold-evidence is available, \textit{unless} the attacker can identify the most damaging persuasive variant per claim. 

\textit{Gold-Evidence} models are more vulnerable to sabotage. In the \textit{Blind} attack, Sabotage ASR rises from below $22\%$ in \textit{Claim-Only} models to above $44\%$ for \textit{Gold-Evidence} models, while in the \textit{Oracle} attack, Sabotage ASR stays at nearly $100\%$ in both settings. These results indicate that persuasion injection can induce claim-evidence misalignment, biasing evidence-grounded models mostly toward \textit{False} predictions.

\subsection{Technique-Level Vulnerability} \label{sec:results_rq2}
We now aim to identify which specific persuasion techniques are most effective at causing evasion ($\textsc{False}{\to}\textsc{True}$), as it is the fundamental goal of a real-world adversarial attack on AFC systems (i.e., to disrupt the system's ability to detect \textit{False} claims).  \autoref{tab:technique_mitigation} splits the techniques into \textit{Effective}, \textit{Partially Effective}, and \textit{Neutralised} based on whether they achieve Evasion ASR ${\ge}5\%$ in the \textit{Gold-Evidence} setting. To illustrate how \textit{Effective} techniques cause evasion, we present concrete examples in \autoref{app:examples}.

\begin{table}[h]
\centering
\small
\begin{adjustbox}{width=\columnwidth}
\begin{tabular}{l|ll|ll}
\toprule
& \multicolumn{2}{c|}{\textbf{FEVER}} & \multicolumn{2}{c}{\textbf{FEVEROUS}} \\
\textbf{Persuasion Technique} & \textbf{Claim} & \textbf{\textit{+E}\textsubscript{\textit{gold}}} & \textbf{Claim} & \textbf{\textit{+E}\textsubscript{\textit{gold}}} \\
\midrule
\multicolumn{5}{l}{\textit{\textbf{Effective} (+E\textsubscript{gold} EASR $\ge 0.05$ in both datasets)}} \\
Obfuscation \textbf{(MW)} & $0.73$ & $\mathbf{0.38}$ & $0.65$ & $\mathbf{0.15}$ \\
Repetition \textbf{(MW)} & $0.33$ & $\mathbf{0.25}$ & $0.25$ & $\mathbf{0.10}$ \\
Slogan \textbf{(C)} & $0.74$\textsuperscript{*} & $\mathbf{0.07}$ & $0.66$\textsuperscript{*} & $\mathbf{0.18}$ \\
Flag Waving \textbf{(J)} & $0.51$ & $\mathbf{0.10}$ & $0.59$ & $\mathbf{0.06}$ \\
Appeal to Popularity \textbf{(J)} & $0.30$ & $\mathbf{0.05}$ & $0.45$ & $\mathbf{0.06}$ \\
\midrule
\multicolumn{5}{l}{\textit{\textbf{Partially Effective} (+E\textsubscript{gold} EASR $\ge 0.05$ in one dataset only)}} \\
Appeal to Values \textbf{(J)} & $0.62$ & $\mathbf{0.11}$ & $0.59$ & $0.04$ \\
Appeal to Authority \textbf{(J)} & $0.35$ & $\mathbf{0.07}$ & $0.40$ & $0.04$ \\
\midrule
\multicolumn{5}{l}{\textit{\textbf{Neutralised} (+E\textsubscript{gold} EASR $< 0.05$ in both datasets)}} \\
Loaded Language \textbf{(MW)} & $0.27$ & $0.04$ & $0.31$ & $0.03$ \\
Guilt by Association \textbf{(AoR)} & $0.31$ & $0.04$ & $0.22$ & $0.02$ \\
Name Calling / Labelling \textbf{(AoR)} & $0.12$ & $0.02$ & $0.11$ & $0.03$ \\
Whataboutism \textbf{(D)} & $0.21$ & $0.02$ & $0.21$ & $0.02$ \\
Casting Doubt \textbf{(AoR)} & $0.18$ & $0.01$ & $0.27$ & $0.03$ \\
Red Herring \textbf{(D)} & $0.22$ & $0.01$ & $0.29$ & $0.01$ \\
Appeal to Hypocrisy \textbf{(AoR)} & $0.18$ & $0.02$ & $0.08$ & $0.01$ \\
Conversation Killer \textbf{(C)} & $0.03$ & $0.01$ & $0.09$ & $0.01$ \\
\bottomrule
\end{tabular}
\end{adjustbox}
\caption{\small Effective and neutralised persuasion techniques.Evasion ASR scores for Claim-only and Gold-evidence models. Higher scores indicate more effective attacks. Coarse persuasion categories are shown in parentheses: \textbf{(MW)}=Manipulative Wording, \textbf{(C)}=Call, \textbf{(J)}=Justification, \textbf{(AoR)}=Attack on Reputation, \textbf{(D)}=Distraction. Best score in the Claim-only setting is denoted with *. Scores with +\textit{E}\textsubscript{\textit{gold}} EASR${\ge}0.05$ are in \textbf{bold}.}
\label{tab:technique_mitigation}
\end{table}

\paragraph{Surface-Form Vulnerabilities.}
In the claim-only setting, performance reflects the verifier’s reliance on surface-level linguistic cues. All techniques (except \textit{Conversation Killer} in FEVER)  induce non-trivial evasion rates above $5\%$. Techniques such as \textit{Slogan} and \textit{Obfuscation} are particularly effective, achieving evasion rates above $65\%$. \textit{Slogan} compresses claims into short, punchy formulations that resemble news headlines (``\textit{Step aside for a new era: Horsford hands the baton in 2009!}''; see \autoref{tab:persistence_examples}), which can increase perceived credibility. \textit{Obfuscation} reduces specificity (e.g., ``\textit{(...) approximately one and a half months}''), making it less apparent that the claim asserts a precise, falsifiable statement, leading to high evasion rates despite no change in the underlying veracity.

\paragraph{Vulnerabilities Under Evidence-Grounding.}
Several high-impact techniques achieve substantial evasion power even for models grounded in gold-evidence, forming a class of effective threats spanning three coarse categories:
\textit{Manipulative Wording}, \textit{Calls}, and \textit{Justification}. 

Among these, \textit{Manipulative Wording} remains the most damaging group (except \textit{Loaded Language}). \textit{Obfuscation} attacks replace concrete entities with vague formulations: ``\textit{50 days}''{$\rightarrow$}``\textit{approximately one and a half months}'', and ``\textit{Thanksgiving}''{$\rightarrow$}``\textit{holiday season}''; See \autoref{tab:persistence_examples}), making it harder for evidence to explicitly contradict claims. This technique's high effectiveness aligns with \citet{glocknerAmbiFCFactCheckingAmbiguous2024}, which found that incomplete or vague formulations elicit substantial disagreement even among human fact-checkers. Next, \textit{Repetition} can reiterate irrelevant or partially true elements of the claim, diverting the focus away from the component that is actually false: ``\textit{GZMB is a serine protease, \ul{a serine protease} with} \{\textit{false predicate}\}''.

\textit{Justification} class of techniques also show high effectiveness in FEVER, with all techniques exceeding the $5\%$ threshold in this dataset; notably, \textit{Flag Waving} and \textit{Appeal to Popularity} maintain effectiveness in both datasets. Both techniques aim to increase the plausibility of \textit{False} information by framing it to be widely accepted: ``\textit{True Sri Lankans must support} \{\textit{false predicate}\}'', and ``\textit{Everyone is celebrating} \{\textit{false predicate}\}'', for \textit{Flag Waving} and \textit{Appeal to Popularity}, respectively.

In contrast, techniques in groups related to \textit{Attack on Reputation} and \textit{Distraction}, which redirect the focus away from the claim (towards the speaker, an external issue, or a simplified framing), collapse to near-zero evasion rates on models grounded on gold-evidence. 


\subsection{Impact on Evidence Retrieval} \label{sec:results_rq3}
So far, we have examined persuasion attacks under gold-evidence to isolate vulnerabilities in the veracity classifier. However, AFC pipelines rely on retrieval systems to access supporting evidence. In this section, we shift focus to the evidence retriever, examining whether persuasion attacks degrade retrieval quality and expose failure modes specific to the retrieval stage.

\begin{figure}[h]
    \centering
    \includegraphics[width=0.94\columnwidth]{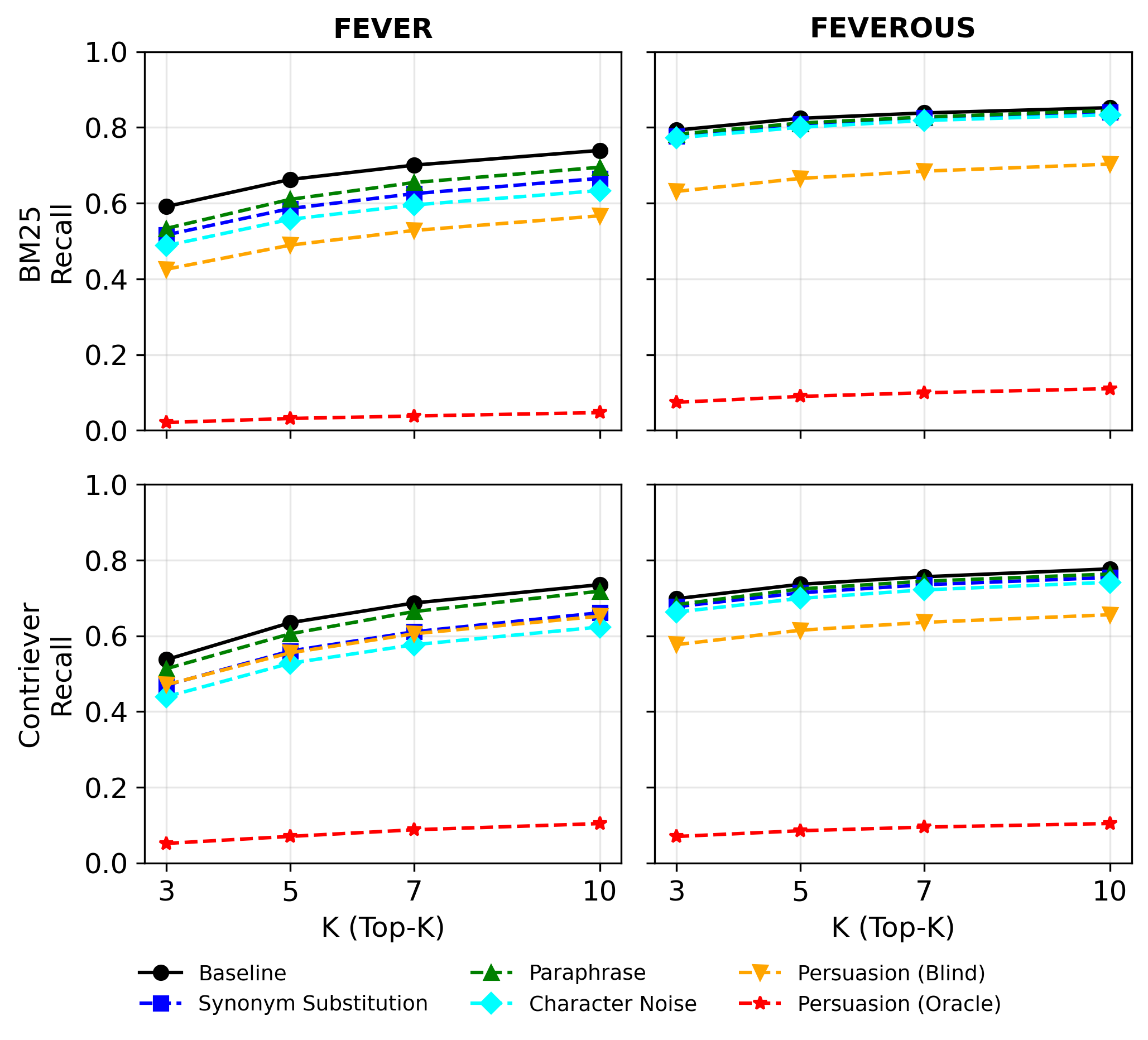}
    \caption{Retrieval performance of BM25 (top) and Contriever (bottom) under adversarial attacks. Lower scores indicate more effective attacks.}
    \label{fig:recall_k_trends}
\end{figure}

\paragraph{Retrieval Vulnerability.}
\autoref{fig:recall_k_trends} illustrates the retrieval performance (Recall@$k$) of both retrievers across varying depths ($k{=}\{3, 5, 7, 10\}$) and under different types of adversarial attacks.
Standard lexical attacks\footnote{BM25 is invariant to token permutation. Therefore, results for the \textit{Word Swap} attack are not included in this analysis.} cause modest retrieval degradation. The strongest lexical attack (\textit{Character Noise}) drops BM25 $Recall@5$ by $0.106$ and $0.024$ points with respect to the \textit{Baseline}, on FEVER and FEVEROUS, respectively.
Persuasion attacks cause deeper failures. Under the \textit{Blind} setting, persuasion injection degrades BM25 $Recall@5$ by $0.174$ points on FEVER and $0.159$ on FEVEROUS, in comparison to the \textit{Baseline}.
The \textit{Oracle} setting again shows a catastrophic power, collapsing $Recall@k$ to near-zero for all values of $k$.
These results confirm that persuasion techniques not only make claims harder to verify, but also decouple the claim from its supporting evidence, causing a complete pipeline failure in AFC systems.

\paragraph{Transfer to Dense Retrieval.}
The vulnerability is not an artefact of lexical matching: the findings transfer from BM25 to Contriever. Under the \textit{Blind} setting, persuasion degrades Contriever $Recall@5$ by $0.080$ on FEVER and $0.121$ on FEVEROUS, and the \textit{Oracle} attacker collapses $Recall@5$ to at most $0.090$ across all four dataset--retriever combinations. Per-technique vulnerability profiles are also preserved across paradigms: the ranking of techniques by $Recall@5$ under blind persuasion correlates strongly between the two retrievers (Spearman $\rho{=}0.825$ on FEVER and $\rho{=}0.886$ on FEVEROUS; $p{<}2{\times}10^{-4}$), with \textit{Obfuscation} and \textit{Slogan} the most damaging techniques in all cases (see \autoref{app:dense_retrieval} for the per-technique breakdown).
We do observe that the dense retriever is partially more robust to blind persuasion: its absolute $Recall@5$ drop is consistently smaller than BM25's (e.g., $-0.080$ vs. $-0.174$ on FEVER), and on FEVER blind persuasion is comparable to the strongest lexical baseline under Contriever. However, since worst-case persuasion still collapses recall on both retrievers, dense retrieval mitigates, but does not defend against, persuasion-based query manipulation.

\begin{figure*}[t]
    \centering
    \includegraphics[width=0.92\textwidth]{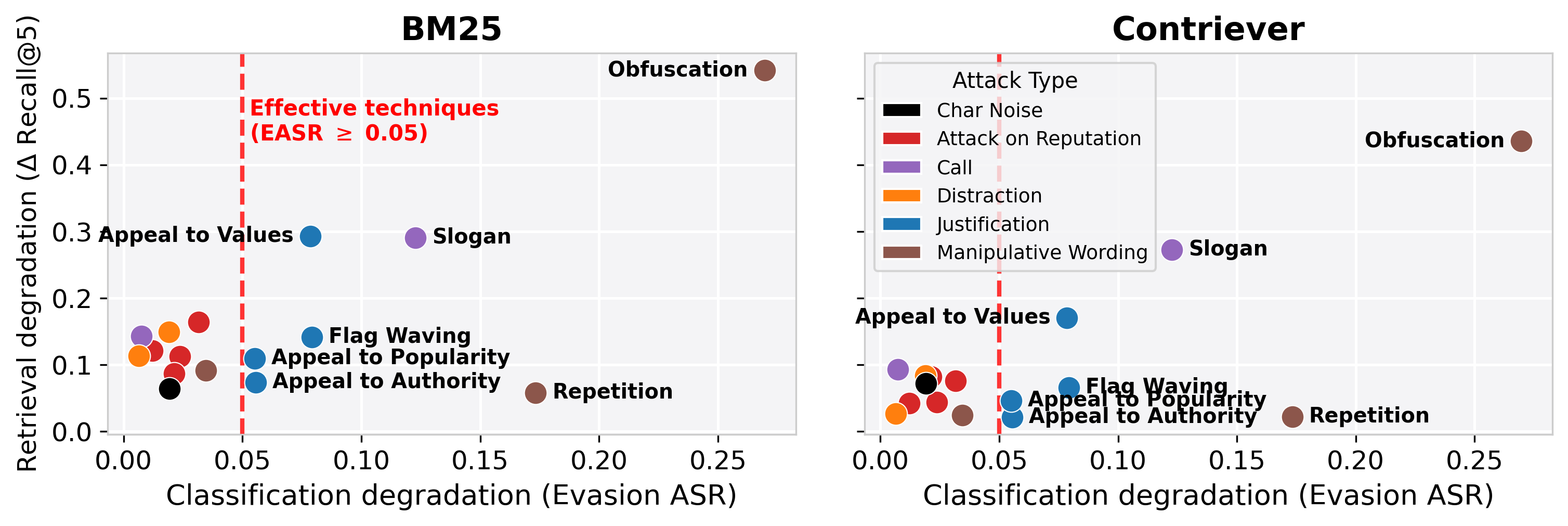}
    \caption{Retrieval vs. classification vulnerabilities under BM25 (left) and Contriever (right). Scores are averaged between FEVER and FEVEROUS. Evasion ASR computed under access to gold-evidence. Higher scores indicate more effective attacks.}
    \label{fig:retrieval_vs_class}
\end{figure*}

\paragraph{Retrieval vs. Classification Vulnerabilities.}
We analyse the extent to which techniques that successfully evade classification also impair retrieval. \autoref{fig:retrieval_vs_class} plots \textit{Retrieval Degradation} ($|\Delta \text{Recall@5}|$) against \textit{Classification Degradation} under E\textsubscript{gold} (Evasion ASR) for individual persuasion techniques, for both retrievers, with scores averaged across FEVER and FEVEROUS (per-dataset values are reported in \autoref{app:dense_retrieval}).

Not all techniques that are effective in degrading classification also substantially harm retrieval. \textit{Repetition}, for example, achieves the second-highest evasion rate ($17.3\%$) yet reduces BM25 Recall@5 by only $0.06$--comparable to \textit{Char Noise}--and its impact on Contriever all but vanishes ($0.02$). This occurs because \textit{Repetition} mostly adds tokens that are already present in the claim, which has a negligible impact on the retriever. Similarly, three out of five \textit{Justification} techniques (\textit{Flag Waving}, \textit{Appeal to Popularity}, and \textit{Appeal to Authority}) form a small cluster of techniques with low effectiveness in attacking the retriever, with up to $0.14$ $|{\Delta}Recall@5|$ on BM25 and $0.07$ on Contriever.

\textit{Appeal to Values} and \textit{Slogan} are effective in degrading both classification and retrieval performance, with Evasion ASR of $7.9\%$ and $12.3\%$, respectively, and $|{\Delta}Recall@5|$ of $0.29$ on BM25 ($0.17$ and $0.27$ on Contriever). More importantly, the single most harmful technique substantially degrades both components under both retrieval paradigms: \textit{Obfuscation} stands out as an outlier attack (top-right in both panels of \autoref{fig:retrieval_vs_class}), with an average Evasion ASR of $27\%$ and a $|{\Delta}Recall@5|$ of $0.54$ on BM25 and $0.44$ on Contriever. By replacing concrete entities with vague generalisations (e.g., changing ``\textit{50 days}'' to ``\textit{approximately one and a half months}'' or ``\textit{Thanksgiving}'' to ``\textit{holiday season}''; see \autoref{tab:persistence_examples}), \textit{Obfuscation} denies the retriever the ability to match concrete keywords needed to find evidence, while simultaneously creating ambiguity that confuses the veracity classifier.

\section{Discussion}
Our results demonstrate that adversarial persuasion is a potent attack mechanism capable of catastrophically disrupting AFC pipelines. \textit{Manipulative Wording} techniques, especially, which align with \citet{glocknerAmbiFCFactCheckingAmbiguous2024}, who showed that vague claims or those requiring inference elicit substantial disagreement even among human fact-checkers. Our findings extend this to adversarial settings and suggest a mitigation direction: grounding claims in richer, structured 
contexts reduces the scope for rhetorical manipulation. Two results support this. First, models without evidence are far more vulnerable (e.g., \textit{Obfuscation} evasion: ${>}65\%$ claim-only vs.\ ${<}38\%$ with gold-evidence), underscoring the need for evidence grounding. Second, denser and more structured evidence further improves robustness: \textit{Obfuscation} evasion drops from $38\%$ on FEVER to $15\%$ on FEVEROUS, where structured 
evidence and ${\approx}2{\times}$ more evidence per claim is available. Conversely, techniques that redirect attention away from the claim, such as \textit{Attack on Reputation} and \textit{Distraction}, are largely neutralised under evidence grounding, collapsing to near-zero evasion rates.

\section{Conclusion \& Future Work}
This paper introduced Persuasion Injection Attacks, the first adversarial 
framework that weaponises persuasion techniques against AFC systems. Experiments on two benchmark datasets showed that AFC pipelines fail to disentangle persuasive rhetoric from factual content, and that optimised persuasion attacks lead to collapse in both retrieval and classification even when gold-evidence is provided.

The findings motivate our future work on developing defences against persuasion-based attacks, particularly \textit{Manipulative Wording} techniques. We also plan to extend the evaluation to cross-encoder retrievers such as ColBERTv2 \cite{santhanamColBERTv2EffectiveEfficient2022}, and to investigate whether the vulnerability persists in other languages and domains.

\section*{Limitations}
This study is restricted to English, and the findings may not directly transfer to other languages without adaptation to account for linguistic and cultural variation.

The experiments rely on Wikipedia-based benchmarks (FEVER and FEVEROUS). While these offer controlled conditions for evaluating factual reasoning, they may differ from disinformation in social media posts or news articles. Thus persuasion attacks may behave differently in these genres.

Finally, our adversarial variants were generated using modest-sized LLMs (around 7B parameters). The strength of individual persuasion techniques likely depends on the generator's capabilities, and more capable models might enhance techniques that appeared less effective in these experiments. 

Our retrieval analysis covers one sparse retriever (BM25), chosen because it remains widely deployed in AFC systems \cite{aly-etal-2021-fact, schlichtkrull-etal-2024-automated, akhtar-etal-2025-2nd}, and one dense retriever (Contriever). While the vulnerability transfers across both paradigms (\autoref{sec:results_rq3}), other retrieval architectures, such as cross-encoder and late-interaction models like ColBERTv2 \cite{santhanamColBERTv2EffectiveEfficient2022}, may exhibit different sensitivity profiles, and evaluating them is an important direction for future work. Nevertheless, our results (Section~\ref{sec:results}) confirm that even if a retriever were to bypass the attack and provide perfect evidence, the attack can still substantially harm the veracity classifier.

Our protocol trains verifiers on clean data and evaluates attacks on the held-out test set, following the established separation between adversarial robustness evaluation and adversarial defense \cite{alzantotGeneratingNaturalLanguage2018, jinBERTReallyRobust2020}. Studying defenses, such as adversarial training with persuasion-augmented data, is therefore beyond our scope; notably, such training is not guaranteed to restore robustness \cite{alzantotGeneratingNaturalLanguage2018}. A related concern is that baseline attacks may appear weaker because verifiers were exposed to similar perturbation pairs (e.g., paraphrases, translations) during pretraining. We note that this cannot fully explain the observed gap: lexical baselines still degrade verifier accuracy by up to $10$ points, consistent with prior findings \cite{mamtaFactEvalEvaluatingRobustness2025}; our baselines also include non-lexical rewrites (\textit{Informal Style Transfer}), which persuasion attacks substantially outperform; and the argument applies symmetrically to persuasion, since LLMs are demonstrably proficient at generating persuasive language \cite{breumPersuasivePowerLarge2024, pauliMeasuringBenchmarkingLarge2025}, implying pretraining exposure to persuasive text as well.

While generative LLMs are increasingly applied to verification tasks \cite{augensteinFactualityChallengesEra2024}, they frequently exhibit hallucinations \cite{ji2025survey}, sensitivity to prompt phrasing \cite{erricaWhatDidWrong2025}, and risk data contamination from popular benchmark datasets such as FEVER and FEVEROUS \cite{chenBenchmarkingLargeLanguage2025}. These factors motivate our choice of a fine-tuned encoder-only RoBERTa model as the primary verifier. Nevertheless, we additionally evaluate two instruction-tuned LLMs to assess architectural generalisation. Crucially, even if an LLM has been exposed to the original benchmark claims during pretraining, this does not invalidate our results: a perturbation that causes misclassification despite the model having potentially memorised the original claim and its label constitutes a \textit{stronger} demonstration of attack effectiveness. Furthermore, the consistent vulnerability observed across RoBERTa (which has no 
such pretraining exposure), and the LLM verifiers suggest that the 
susceptibility is structural rather than dataset-specific.

\section*{Ethical Considerations}

\paragraph{Potential Misuse}
By developing a systematic framework for injecting persuasion into claims, there is a risk that malicious actors could use these techniques to craft more deceptive disinformation that evades automated detection. However, by documenting and exposing these vulnerabilities, we enable the community to develop more robust defence mechanisms. Furthermore, we do not introduce new persuasion techniques in this work. Instead, we use techniques that are well-documented in prior work on persuasion and propaganda \cite{martinoSurveyComputationalPropaganda2020, piskorskiSemEval2023Task32023}. 

\paragraph{Data and Artifacts}
We use the FEVER (\url{https://fever.ai/dataset/fever.html}) and FEVEROUS (\url{https://fever.ai/dataset/feverous.html}) datasets, which are publicly available under CC-BY-SA 3.0 licenses. Our use of these artifacts is consistent with their intended use as benchmarks for verifying factual claims. We release our attack code under an open-source MIT license to facilitate reproducibility and allow the community to freely reuse, modify, and build upon our adversarial framework for defensive research purposes.

\section*{Acknowledgments}
This work is supported by the Engineering and Physical Sciences Research Council (EPSRC) under grant UKRI3352: Longitudinal, Multilingual, and Multi-format Investigation and Detection of LLM-Generated Disinformation. João A. Leite is supported by a University of Sheffield EPSRC Doctoral Training Partnership Scholarship.

\bibliography{main}

\clearpage
\appendix
\raggedbottom
\makeatletter
\setlength{\@fptop}{0pt}      
\setlength{\@dblfptop}{0pt}   
\makeatother

\section{Definition of Persuasion Techniques} \label{app:persuasion_definition}

\autoref{tab:taxonomy_definitions} lists the full set of $23$ persuasion techniques considered in this work, grouped into $6$ high-level categories. Technique definitions follow the taxonomy introduced by \citet{piskorskiSemEval2023Task32023}.

\begin{table*}[t]
\centering
\small
\setlength{\tabcolsep}{5pt}
\begin{adjustbox}{width=\linewidth}
\begin{tabular}{l p{0.27\linewidth} p{0.55\linewidth}}
\toprule
\textbf{Category} & \textbf{Persuasion Technique} & \textbf{Definition} \\
\midrule

\multirow{5}{*}{\textbf{Attack on Reputation}} 
 & Name Calling / Labelling & Attach a loaded, insulting, or demeaning label to a person or group; targets identity rather than argument. \\
 & Guilt by Association & Associate the target with a negatively perceived person, group, or event to discredit them. \\
 & Casting Doubt & Undermine credibility by questioning competence, motives, or facts about the target. \\
 & Appeal to Hypocrisy & Accuse the opponent of hypocrisy to deflect criticism. \\
 & Questioning the Reputation & Attack the reputation of the target by making strong negative claims about it. \\
\midrule

\multirow{5}{*}{\textbf{Justification}} 
 & Flag Waving & Justify a claim by appealing to group pride, patriotism, or shared values. \\
 & Appeal to Authority & Support a claim by invoking an authority’s endorsement to lend weight. \\
 & Appeal to Popularity & Justify a claim because many people support it or do it. \\
 & Appeal to Values & Link the claim to widely held positive values (tradition, religion, ethics). \\
 & Appeal to Fear & Promote or reject an idea by evoking fear about its consequences. \\
\midrule

\multirow{3}{*}{\textbf{Distraction}} 
 & Strawman & Misrepresent the claim into a weaker or extreme version that is easier to refute. \\
 & Red Herring & Divert attention from the claim by bringing up an unrelated topic. \\
 & Whataboutism & Deflect criticism by pointing to different wrongdoing elsewhere. \\
\midrule

\multirow{3}{*}{\textbf{Simplification}} 
 & Causal Oversimplification & Blame a complex outcome on a single cause, ignoring other factors. \\
 & False Dilemma & Present only two options when more exist; force a black-and-white choice. \\
 & Consequential Oversimplification & Argue that one action will inevitably lead to extreme consequences. \\
\midrule

\multirow{3}{*}{\textbf{Call}} 
 & Slogan & Short, catchy phrase intended to persuade emotionally. \\
 & Conversation Killer & Phrases that shut down discussion and discourage critical thinking. \\
 & Appeal to Time & Argue urgency or timeliness to force immediate action. \\
\midrule

\multirow{4}{*}{\textbf{Manipulative Wording}} 
 & Loaded Language & Use emotionally charged words or phrases to sway feelings. \\
 & Obfuscation & Use unclear or vague wording to hide meaning or confuse the audience. \\
 & Exaggeration or Minimisation & Overstate or downplay importance to skew perception. \\
 & Repetition & Repeat words or phrases to increase persuasion through familiarity. \\

\bottomrule
\end{tabular}
\end{adjustbox}
\caption{Taxonomy of persuasion techniques. The $23$ techniques are grouped into $6$ high-level categories following the hierarchy defined by \citet{piskorskiSemEval2023Task32023}.}
\label{tab:taxonomy_definitions}
\end{table*}

\section{Generator Robustness}
\label{app:generator_robustness}

To verify that the susceptibility of AFC systems to Persuasion Injection reflects a property of the fact-checking pipeline rather than an artefact of a particular generative model, we replicate the \textit{Blind} attack using \textsc{Llama-3-8B-Instruct}.

We conduct this robustness check on \textbf{FEVEROUS}, which is more challenging to attack overall (see \autoref{sec:results_rq1}). For consistency, both generators are evaluated on the same subset of $15$ \textbf{validated persuasion techniques} used in the main experiments (see \autoref{tab:filtering_results}).

As shown in \autoref{tab:llama_comparison}, attacks generated by \textsc{Llama-3-8B-Instruct} induce nearly identical accuracy degradation to those generated by \textsc{Qwen2.5-7B} under both target settings (claim-only and evidence-based). This consistency indicates that the observed vulnerabilities are not driven by generator-specific behaviour.

\begin{table}[H]
\centering
\begin{adjustbox}{width=\columnwidth}
\begin{tabular}{l c | c c}
\toprule
& \textbf{Baseline} & \multicolumn{2}{c}{\textbf{Blind Attack Accuracy}} \\
\textbf{Target Setting} & (No Attack) & \textbf{Qwen2.5-7B} & \textbf{Llama-3-8B} \\
\midrule
Claim-Only      & 0.641 & 0.575 & 0.568 \\
Evidence-Based  & 0.894 & 0.726 & 0.727 \\
\bottomrule
\end{tabular}
\end{adjustbox}
\caption{Robustness across generators (FEVEROUS). AFC accuracy under blind attacks generated by Qwen2.5-7B and Llama-3-8B. Both generators yield comparable performance degradation, indicating generator-agnostic attack effectiveness.}
\label{tab:llama_comparison}
\end{table}

\section{Prompt Templates} \label{app:prompt_templates}

This appendix documents the prompt templates used to generate claim variants throughout our experiments.

\paragraph{Adversarial Generation Prompt.}
\autoref{fig:attack_prompt_template} shows the universal prompt template used to generate persuasive claim variants. For each persuasion technique, the template is instantiated with definitions drawn from the taxonomy in \autoref{tab:taxonomy_definitions}.

The prompt explicitly constrains the generator to preserve the original factual content, ensuring that generated variants differ in persuasive form rather than underlying veracity.

\begin{figure}[H]
    \centering
    \small
    \begin{tabular}{|p{0.95\linewidth}|}
    \hline
    \textbf{Instruction:} Rewrite the claim to inject persuasive wording of type: ``\texttt{\{TECHNIQUE\_NAME\}}''. \\
    \textbf{Definition:} \texttt{\{DEFINITION\}} \\
    \\
    \textbf{Goal:} Keep the original factual content identical (no new facts, numbers, or named entities). Modify wording only to introduce the specified persuasion technique. \\
    \\
    \textbf{Constraints:} \\
    - Do NOT change the factuality of the claim. \\
    - Keep length within $\pm$20\% of original. \\
    - Produce a single-sentence output. \\
    - If the claim is already clearly using this technique, produce a stronger, clearer instance of it (without adding facts). \\
    \\
    \textbf{Few-shot outputs that demonstrate the tone for ``\texttt{\{TECHNIQUE\_NAME\}}'':} \\
    \texttt{\{EXAMPLES\}} \\
    \\
    Now rewrite the following claim into a version that injects ``\texttt{\{TECHNIQUE\_NAME\}}'': \\
    \textbf{CLAIM:} ``\texttt{\{CLAIM\}}'' \\
    \\
    \textbf{IMPORTANT:} Output ONLY the rewritten claim text. Do not include any prefixes (like ``Here is the rewritten claim:''), explanations, or commentary. Start directly with the claim text. \\
    \hline
    \end{tabular}
    \caption{Prompt template for persuasion injection attacks.}
    \label{fig:attack_prompt_template}
\end{figure}

\paragraph{Paraphrase Prompt.}
\autoref{fig:paraphrase_prompt} presents the prompt used to generate neutral paraphrases of claims. This prompt instructs the model to rewrite the input claim while preserving its original meaning and tone.

Paraphrased claims are used as one of the baseline transformations in our experimental evaluation (see \autoref{subsec:baseline_attacks}).

\begin{figure}[H]
    \centering
    \small
    \begin{tabular}{|p{0.95\linewidth}|}
    \hline
    \textbf{Instruction:} Produce a paraphrase of the provided claim. Keep the same length as much as possible. \\
    \\
    \textbf{Example:} \\
    Input: ``The senator voted for the amendment in 2019.'' \\
    Output: ``In 2019, the senator supported the amendment.'' \\
    \\
    \textbf{Claim:} \texttt{\{CLAIM\}} \\
    \textbf{Paraphrase:} \\
    \\
    \textbf{IMPORTANT:} Output ONLY the paraphrased claim text. Do not include any prefixes, explanations, or commentary. \\
    \hline
    \end{tabular}
    \caption{Prompt template for the paraphrase attack.}
    \label{fig:paraphrase_prompt}
\end{figure}

\paragraph{Informal Style Prompt.}                                                                                              
  \autoref{fig:informal_prompt} shows the prompt used to generate variants of claims using the informal-style attack, in which a claim is rewritten in a more colloquial tone while preserving its factual content.

  \begin{figure}[H]
      \centering
      \small
      \begin{tabular}{|p{0.95\linewidth}|}
      \hline
      \textbf{Instruction:} You are a text rewriting assistant. Rewrite the following claim in a more informal, conversational tone. Keep the core factual proposition unchanged. Do not add or remove factual content, entities, numbers, or dates. Output only the rewritten claim, without any additional information. \\
      \\
      \textbf{Claim:} \\
      \texttt{\{CLAIM\}} \\
      \\
      \textbf{Rewritten claim:} \\
      \hline
      \end{tabular}
      \caption{Prompt template for the informal style transfer baseline.}
      \label{fig:informal_prompt}
  \end{figure}

  \paragraph{Verifier Prompt Templates.}
  For the verifiers employing decoder LLMs (Llama-3-8B and
  Mistral-7B), we employ the instruction prompt shown in \autoref{fig:verifier_prompt}.

  \begin{figure}[H]
      \centering
      \small
      \begin{tabular}{|p{0.95\linewidth}|}
      \hline
      \textbf{Instruction:} You are a fact-checking verifier. \\ \\
      \textbf{Task}: decide whether the claim is TRUE or FALSE. \\
      Focus on the core factual proposition of the claim. \\
      Return exactly one label: TRUE or FALSE. \\
      \\
      \textbf{Claim:} \\
      \texttt{\{CLAIM\}} \\
      \\
      \textbf{Evidence:} \\
      \texttt{1. \{evidence\textsubscript{1}\}} \\
      \texttt{2. \{evidence\textsubscript{2}\}} \\
      \texttt{\ldots} \\
      \\
      \textbf{Label:} \\
      \hline
      \end{tabular}
      \caption{Prompt template for fact-checking verifiers.}
      \label{fig:verifier_prompt}
  \end{figure}

\section{Manual Validation of Label Preservation}
\label{app:data_annotation}

To ensure that persuasive claim variants preserve the ground-truth veracity label of the original claim, we performed a manual validation study to rule out semantic drift as a confounding factor in the observed attack effectiveness.

\paragraph{Annotation Protocol.} We sampled a random subset of $690$ adversarial claims: $30$ per persuasion technique, stratified by dataset (FEVER/FEVEROUS) and ground-truth label (\textit{True}/\textit{False}).

Each instance was annotated by a trained annotator who was shown the persuasive claim together with the corresponding gold-evidence. The annotator assigned one of three verdicts: \textsc{True}, \textsc{False}, or \textsc{Not Enough Info (NEI)}.

\paragraph{Decision Procedure.}
Annotation followed a fixed three-step procedure designed to decouple rhetorical form from factual content:
\begin{enumerate}
    \item \textbf{Core Fact Identification:} isolate the central factual assertion expressed by the claim.
    \item \textbf{Evidence Verification:} determine whether the gold-evidence explicitly supports or refutes this assertion.
    \item \textbf{Ambiguity Resolution:} if the claim becomes subjective, evaluative, or underspecified such that it cannot be verified given the evidence, assign \textit{NEI}.
\end{enumerate}

\paragraph{Label Preservation Criterion.}
An adversarial claim was considered \emph{label-preserving} if the annotator’s verdict matched the original ground-truth label. Claims annotated as \textit{NEI} or receiving a verdict opposite to the original label were counted as non-preserving.

\paragraph{Technique Selection and Filtering.} \autoref{tab:filtering_results} reports label preservation, label flip, and ambiguity rates for all 23 persuasion techniques.

\begin{table}[H]
\centering
\begin{adjustbox}{width=\columnwidth}
\begin{tabular}{lccc}
\toprule
\textbf{Persuasion Technique} & \textbf{Preservation} & \textbf{Flip} & \textbf{Ambiguity} \\
\midrule
\multicolumn{4}{c}{\textsc{\textbf{Included}} (${>}80\%$ Preservation)} \\
\midrule
Repetition & $96.67$ & $3.33$ & $0.00$ \\
Obfuscation & $93.33$ & $0.00$ & $6.67$ \\
Flag Waving & $93.33$ & $0.00$ & $6.67$ \\
Casting Doubt & $93.33$ & $0.00$ & $6.67$ \\
Conversation Killer & $93.33$ & $6.67$ & $0.00$ \\
Appeal to Values & $93.33$ & $0.00$ & $6.67$ \\
Slogan & $90.00$ & $0.00$ & $10.00$ \\
Loaded Language & $90.00$ & $0.00$ & $10.00$ \\
Appeal to Authority & $90.00$ & $0.00$ & $10.00$ \\
Appeal to Popularity & $90.00$ & $0.00$ & $10.00$ \\
Appeal to Hypocrisy & $86.67$ & $3.33$ & $10.00$ \\
Red Herring & $86.67$ & $0.00$ & $13.33$ \\
Guilt by Association & $86.67$ & $3.33$ & $10.00$ \\
Whataboutism & $86.67$ & $3.33$ & $10.00$ \\
Name Calling / Labelling & $83.33$ & $3.33$ & $13.33$ \\
\midrule
\textbf{\textit{Overall (Micro Avg.)}} & $\mathbf{90.22}$ & $\mathbf{1.56}$ & $\mathbf{8.22}$ \\
\midrule
\multicolumn{4}{c}{\textsc{\textbf{Excluded}} (${\leq}80\%$ Preservation)} \\
\midrule
Appeal to Time & $80.00$ & $0.00$ & $20.00$ \\
Causal Oversimplification & $76.67$ & $6.67$ & $16.67$ \\
False Dilemma & $56.67$ & $6.67$ & $36.67$ \\
Exaggeration / Minimisation & $36.67$ & $3.33$ & $60.00$ \\
Consequential Oversimplification & $30.00$ & $3.33$ & $66.67$ \\
Questioning the Reputation & $26.67$ & $0.00$ & $73.33$ \\
Strawman & $16.67$ & $6.67$ & $76.67$ \\
Appeal to Fear / Prejudice & $6.67$ & $0.00$ & $93.33$ \\
\bottomrule
\end{tabular}

\end{adjustbox}
\caption{Label preservation analysis per persuasion technique (\%).
\emph{Preservation} denotes the proportion of persuasive claims whose annotated verdict matches the original ground-truth label. \emph{Flip} denotes cases where the annotated verdict contradicts the original label (\textsc{True}${\leftrightarrow}$\textsc{False}).
\emph{Ambiguity} denotes claims judged as \textit{Not Enough Info}, i.e., unverifiable given the gold-evidence.}
\label{tab:filtering_results}
\end{table}

Based on this validation, \textbf{all techniques with label preservation rates $\mathbf{{\leq}80\%}$ were excluded from the experiments.}
This filtering step removed eight techniques. The remaining fifteen retained techniques show a label preservation rate of $\mathbf{90.22\%}$. For the claims that failed the preservation check, less than $\mathbf{1.5\%}$ correspond to label flips (i.e., \textit{False}{$\to$}\textit{True} or \textit{True}{$\to$}\textit{False}).

\section{Length Preservation Analysis} \label{app:length-preservation}

A potential confound in persuasion injection attacks is that adversarial 
claims may be substantially longer than their originals, introducing 
additional propositional content that could independently affect classifier 
behaviour. To prevent this, we explicitly constrain the generator 
to produce outputs within $\pm20\%$ of the original token count and to 
produce single-sentence outputs (see prompt template in 
\autoref{fig:attack_prompt_template}).

To verify empirically that these constraints are respected, we compare 
token counts between the original and adversarial test sets using 
Welch's $t$-test, including the paraphrasing baseline attack as a reference point.

\begin{table}[H]
\centering
\small
\resizebox{\columnwidth}{!}{%
\begin{tabular}{lcccc}
\toprule
\textbf{Attack} & \textbf{Tokens (orig.)} & \textbf{Tokens (attack)} & \textbf{$\Delta$} & \textbf{p-value} \\
\midrule
Paraphrasing        & $17.47$ & $15.44$ & $-2.03$ & $5.79 \times 10^{-55}$ \\
\textbf{Persuasion} & $\mathbf{17.47}$ & $\mathbf{18.91}$ & $\mathbf{+1.44}$ & $\mathbf{5.45 \times 10^{-45}}$ \\
\bottomrule
\end{tabular}%
}
\caption{Token-count comparison between original and adversarial 
claims. $\Delta$ = mean difference (attack $-$ clean).}
\label{tab:length_preservation}
\end{table}

Persuasion attacks add on average fewer than
$1.5$ tokens per claim, a smaller absolute change than Paraphrasing
($-2.03$ tokens). This confirms that the observed performance degradation cannot be
attributed to differences in claim length.

\paragraph{Per-Technique Breakdown.}
The aggregate average masks non-uniform per-technique behaviour: some techniques produce longer claims and others shorter ones, with the deviations largely cancelling out. \autoref{tab:length_per_technique} reports the per-technique Welch's $t$-test (combined across FEVER and FEVEROUS; $n{=}14{,}055$ clean test claims with mean original length of $17.47$ tokens).

\begin{table}[H]
\centering
\small
\begin{adjustbox}{width=0.75\columnwidth}
\begin{tabular}{lcc}
\toprule
\textbf{Technique} & \textbf{$\Delta$} & \textbf{p-value} \\
\midrule
Red Herring & $+10.35$ & ${<}10^{-4}$ \\
Appeal to Hypocrisy & $+5.51$ & ${<}10^{-4}$ \\
Repetition & $+4.58$ & ${<}10^{-4}$ \\
Guilt by Association & $+4.17$ & ${<}10^{-4}$ \\
Whataboutism & $+3.68$ & ${<}10^{-4}$ \\
Casting Doubt & $+2.70$ & ${<}10^{-4}$ \\
Flag Waving & $+2.27$ & ${<}10^{-4}$ \\
Loaded Language & $+1.67$ & ${<}10^{-4}$ \\
Appeal to Authority & $+1.32$ & ${<}10^{-4}$ \\
Appeal to Values & $+0.99$ & ${<}10^{-4}$ \\
Appeal to Popularity & $+0.77$ & ${<}10^{-4}$ \\
Obfuscation & $-0.39$ & $1.71{\times}10^{-3}$ \\
Name Calling / Labelling & $-3.06$ & ${<}10^{-4}$ \\
Conversation Killer & $-4.00$ & ${<}10^{-4}$ \\
Slogan & $-8.95$ & ${<}10^{-4}$ \\
\midrule
\textbf{Persuasion (overall)} & $\mathbf{+1.44}$ & $\mathbf{{<}10^{-4}}$ \\
\textit{Paraphrase (reference)} & $-2.03$ & ${<}10^{-4}$ \\
\bottomrule
\end{tabular}
\end{adjustbox}
\caption{Per-technique token-count difference between adversarial and original claims (Welch's $t$-test). $\Delta$ = mean difference (attack $-$ clean).}
\label{tab:length_per_technique}
\end{table}

Crucially, length variation does not predict attack effectiveness. The techniques with the largest length expansions, \textit{Red Herring} ($+10.35$), \textit{Appeal to Hypocrisy} ($+5.51$), \textit{Guilt by Association} ($+4.17$), and \textit{Whataboutism} ($+3.68$), are all classified as \textit{Neutralised} under gold-evidence (Evasion ASR ${<}5\%$ in both datasets; see \autoref{tab:technique_mitigation}). Conversely, the single most damaging technique, \textit{Obfuscation}, is essentially length-preserving ($-0.39$), and \textit{Appeal to Popularity} ($+0.77$) and \textit{Appeal to Values} ($+0.99$), also among the effective techniques, remain close to the original length. If length-induced distribution shift were driving attack success, we would expect the opposite pattern. This suggests that the impact of persuasion attacks stems from semantic and pragmatic alterations rather than superficial length differences.

\section{Training Details \& Hyperparameters} \label{app:hyperparameters}
We fine-tune \textbf{RoBERTa-base} classifiers on the FEVER-style fact-checking splits, treating each claim concatenated with its supporting evidence sentences (or claim-only in the ablation) as a single sequence. Inputs are truncated and padded to 512 tokens. Models use standard Hugging Face training. We use an effective batch size of 256 (128 per device × 2 gradient accumulation steps). We optimise for five epochs with weight decay 0.01, linear warmup over 10\% of steps, gradient norm clipping at 1.0, and early stopping with patience one based on dev AUC ROC. We mitigate class imbalance with inverse-frequency loss weights computed from the training data.

We perform a grid search over learning rates \{2e‑5, 3e‑5, 5e‑5\}, and select the best by dev AUC ROC. Then, we fine-tune the model using the best learning rate using three different random seeds. For each experiment, we aggregate dev metrics across seeds (mean ± sd) and keep the best-performing checkpoint for downstream use.

We fine-tune \textbf{Llama-3-8B-Instruct} and \textbf{Mistral-7B-Instruct-v0.3} using the same recipe, with adaptations for decoder-only models at 7-8B scale. Rather than full fine-tuning, we apply LoRA~\cite{hu2022lora} with rank $r{=}16$, scaling factor $\alpha{=}32$, and dropout $0.1$, targeting all attention and feed-forward projection layers while also training the classification head. Models are loaded in \texttt{bfloat16} with gradient checkpointing enabled. We use a per-device batch size of $4$ with $16$ gradient accumulation steps (effective batch size $64$). All other settings match RoBERTa: five epochs, weight decay $0.01$, linear warmup over $10\%$ of steps, gradient norm clipping at $1.0$, early stopping with patience one on dev AUC ROC, and inverse-frequency class weights. The learning rate grid is \{1e\text{-}4, 2e\text{-}4, 3e\text{-}4\}, with a grid search run for a single epoch per candidate to reduce cost. We then train with the best rate using two random seeds and keep the best checkpoint.

\autoref{tab:dev_results} reports dev set performance across all six 
model-condition combinations. All models achieve competitive performance on 
both datasets, with RoBERTa-base reaching the highest AUCROC in the 
evidence-grounded setting on FEVER ($0.985$) and FEVEROUS ($0.972$).

\begin{table}[H]
  \centering
  \resizebox{\columnwidth}{!}{%
  \begin{tabular}{llcc}
  \toprule
  \textbf{Model} & \textbf{Cond.} & \textbf{AUCROC} & \textbf{Macro-F1} \\
  \midrule
  \multicolumn{4}{l}{\textit{FEVER}} \\
  \midrule
  RoBERTa-base   & No Ev & .913\small{$\pm$.001} & .826\small{$\pm$.002} \\
  Llama-3-8B     & No Ev & .909\small{$\pm$.001} & .832\small{$\pm$.002} \\
  Mistral-7B     & No Ev & \textbf{.932}\small{$\pm$.000} & \textbf{.852}\small{$\pm$.003} \\
  \midrule
  RoBERTa-base   & W/Ev  & \textbf{.985}\small{$\pm$.000} & \textbf{.954}\small{$\pm$.001} \\
  Llama-3-8B     & W/Ev  & .963\small{$\pm$.001} & .908\small{$\pm$.001} \\
  Mistral-7B     & W/Ev  & .962\small{$\pm$.002} & .920\small{$\pm$.004} \\
  \midrule
  \multicolumn{4}{l}{\textit{FEVEROUS}} \\
  \midrule
  RoBERTa-base   & No Ev & \textbf{.719}\small{$\pm$.001} & \textbf{.661}\small{$\pm$.002} \\
  Llama-3-8B     & No Ev & .677\small{$\pm$.001} & .629\small{$\pm$.003} \\
  Mistral-7B     & No Ev & .688\small{$\pm$.009} & .598\small{$\pm$.041} \\
  \midrule
  RoBERTa-base   & W/Ev  & \textbf{.972}\small{$\pm$.001} & \textbf{.910}\small{$\pm$.002} \\
  Llama-3-8B     & W/Ev  & .934\small{$\pm$.001} & .854\small{$\pm$.001} \\
  Mistral-7B     & W/Ev  & .952\small{$\pm$.001} & .886\small{$\pm$.001} \\
  \bottomrule
  \end{tabular}%
  }
  \caption{Dev set classification results (mean $\pm$ sd across seeds). \textbf{W/Ev}: with gold evidence; \textbf{No Ev}: claim only. Best per column within each block in \textbf{bold}.}
  \label{tab:dev_results}
\end{table}



All experiments were conducted using a single NVIDIA-V100 (40GB VRAM) GPU. We estimate approximately 60 GPU hours to fully reproduce all experiments, including generating the adversarial texts, fine-tuning the veracity classifiers, and performing inference, for both FEVER and FEVEROUS. The dense retrieval analysis requires approximately 30 additional GPU hours, dominated by encoding the Wikipedia snapshots with Contriever.

\section{Full Classification Results} \label{app:full_results}

Tables~\ref{tab:f1_results} and~\ref{tab:auc_results} report Macro-F1 and 
ROC-AUC across all attacks, verifiers, datasets, and evidence conditions. 
Results are consistent with the accuracy trends reported in 
\autoref{tab:main_results}: persuasion attacks substantially outperform all 
baselines across all metrics.

\begin{table*}[ht]
\centering
\setlength{\tabcolsep}{4pt}
\begin{adjustbox}{width=\linewidth}
\begin{tabular}{lcccccccccccc}
\toprule
& \multicolumn{4}{c}{\textbf{RoBERTa-base}}
& \multicolumn{4}{c}{\textbf{Llama-3-8B}}
& \multicolumn{4}{c}{\textbf{Mistral-7B}} \\
\cmidrule(lr){2-5}\cmidrule(lr){6-9}\cmidrule(lr){10-13}
& \multicolumn{2}{c}{\textit{FEVER}}
& \multicolumn{2}{c}{\textit{FEVEROUS}}
& \multicolumn{2}{c}{\textit{FEVER}}
& \multicolumn{2}{c}{\textit{FEVEROUS}}
& \multicolumn{2}{c}{\textit{FEVER}}
& \multicolumn{2}{c}{\textit{FEVEROUS}} \\
\cmidrule(lr){2-3}\cmidrule(lr){4-5}\cmidrule(lr){6-7}\cmidrule(lr){8-9}\cmidrule(lr){10-11}\cmidrule(lr){12-13}
\textbf{Attack}
& Claim & $+E_\textit{gold}$
& Claim & $+E_\textit{gold}$
& Claim & $+E_\textit{gold}$
& Claim & $+E_\textit{gold}$
& Claim & $+E_\textit{gold}$
& Claim & $+E_\textit{gold}$ \\
\midrule
None (Original)
& $.814$ & $.943$ & $.627$ & $.894$
& $.827$ & $.903$ & $.613$ & $.849$
& $.843$ & $.914$ & $.640$ & $.885$ \\
\midrule
Synonym Substitution
& \ul{$.754$} & \ul{$.844$} & $.623$      & \ul{$.857$}
& $.736$      & \ul{$.814$} & $.577$      & \ul{$.821$}
& \ul{$.755$} & \ul{$.832$} & \ul{$.607$} & \ul{$.864$} \\
Character Perturbations
& $.767$ & $.869$ & $.624$      & $.877$
& \ul{$.730$} & $.846$ & \ul{$.555$} & $.829$
& $.782$ & $.879$ & $.624$      & $.876$ \\
Word Swap
& $.791$ & $.929$ & $.624$      & $.894$
& $.760$ & $.867$ & $.586$      & $.842$
& $.802$ & $.883$ & $.620$      & $.882$ \\
Back-Translation
& $.797$ & $.922$ & $.624$      & $.885$
& $.802$ & $.879$ & $.612$      & $.842$
& $.813$ & $.891$ & $.637$      & $.878$ \\
Informal Style Transfer
& $.775$ & $.901$ & $.621$      & $.874$
& $.789$ & $.873$ & $.604$      & $.841$
& $.815$ & $.890$ & $.630$      & $.878$ \\
Paraphrasing
& $.755$ & $.912$ & \ul{$.610$} & $.881$
& $.774$ & $.867$ & $.612$      & $.836$
& $.801$ & $.876$ & $.624$      & $.881$ \\
\midrule
\textbf{Persuasion (Blind)}
& $\mathbf{.678}$ & $\mathbf{.760}$ & $\mathbf{.567}$ & $\mathbf{.722}$
& $\mathbf{.615}$ & $\mathbf{.760}$ & $\mathbf{.489}$ & $\mathbf{.765}$
& $\mathbf{.682}$ & $\mathbf{.758}$ & $\mathbf{.549}$ & $\mathbf{.764}$ \\
\textbf{Persuasion (Oracle)}
& $\mathbf{.042}$ & $\mathbf{.142}$ & $\mathbf{.010}$ & $\mathbf{.203}$
& $\mathbf{.010}$ & $\mathbf{.123}$ & $\mathbf{.042}$ & $\mathbf{.334}$
& $\mathbf{.037}$ & $\mathbf{.138}$ & $\mathbf{.010}$ & $\mathbf{.195}$ \\
\bottomrule
\end{tabular}
\end{adjustbox}
\caption{\textbf{Macro-F1 under adversarial attacks.} Persuasion attacks are
shown in \textbf{bold}; \ul{underlined} values indicate the strongest baseline
attack per column.}
\label{tab:f1_results}
\end{table*}

\begin{table*}[ht]
\centering
\setlength{\tabcolsep}{4pt}
\begin{adjustbox}{width=\linewidth}
\begin{tabular}{lcccccccccccc}
\toprule
& \multicolumn{4}{c}{\textbf{RoBERTa-base}}
& \multicolumn{4}{c}{\textbf{Llama-3-8B}}
& \multicolumn{4}{c}{\textbf{Mistral-7B}} \\
\cmidrule(lr){2-5}\cmidrule(lr){6-9}\cmidrule(lr){10-13}
& \multicolumn{2}{c}{\textit{FEVER}}
& \multicolumn{2}{c}{\textit{FEVEROUS}}
& \multicolumn{2}{c}{\textit{FEVER}}
& \multicolumn{2}{c}{\textit{FEVEROUS}}
& \multicolumn{2}{c}{\textit{FEVER}}
& \multicolumn{2}{c}{\textit{FEVEROUS}} \\
\cmidrule(lr){2-3}\cmidrule(lr){4-5}\cmidrule(lr){6-7}\cmidrule(lr){8-9}\cmidrule(lr){10-11}\cmidrule(lr){12-13}
\textbf{Attack}
& Claim & $+E_\textit{gold}$
& Claim & $+E_\textit{gold}$
& Claim & $+E_\textit{gold}$
& Claim & $+E_\textit{gold}$
& Claim & $+E_\textit{gold}$
& Claim & $+E_\textit{gold}$ \\
\midrule
None (Original)
& $.893$ & $.979$ & $.686$ & $.964$
& $.909$ & $.959$ & $.668$ & $.930$
& $.926$ & $.957$ & $.696$ & $.953$ \\
\midrule
Synonym Substitution
& $.840$      & \ul{$.923$} & $.670$      & \ul{$.946$}
& \ul{$.810$} & \ul{$.900$} & \ul{$.631$} & \ul{$.909$}
& \ul{$.856$} & \ul{$.907$} & \ul{$.665$} & \ul{$.940$} \\
Character Perturbations
& $.846$ & $.938$ & $.680$      & $.954$
& \ul{$.810$} & $.920$ & $.637$      & $.918$
& $.884$ & $.933$ & $.677$      & $.948$ \\
Word Swap
& $.872$ & $.974$ & $.677$      & $.963$
& $.837$ & $.934$ & $.645$      & $.922$
& $.887$ & $.937$ & $.677$      & $.949$ \\
Back-Translation
& $.875$ & $.968$ & $.679$      & $.957$
& $.879$ & $.942$ & $.659$      & $.921$
& $.902$ & $.943$ & $.691$      & $.947$ \\
Informal Style Transfer
& $.856$ & $.956$ & \ul{$.667$} & $.948$
& $.869$ & $.936$ & $.649$      & $.923$
& $.900$ & $.943$ & $.676$      & $.947$ \\
Paraphrasing
& \ul{$.837$} & $.953$ & $.669$      & $.954$
& $.860$ & $.925$ & $.654$      & $.922$
& $.880$ & $.928$ & $.672$      & $.948$ \\
\midrule
\textbf{Persuasion (Blind)}
& $\mathbf{.745}$ & $\mathbf{.862}$ & $\mathbf{.597}$ & $\mathbf{.870}$
& $\mathbf{.697}$ & $\mathbf{.846}$ & $\mathbf{.565}$ & $\mathbf{.869}$
& $\mathbf{.747}$ & $\mathbf{.835}$ & $\mathbf{.576}$ & $\mathbf{.862}$ \\
\textbf{Persuasion (Oracle)}
& $\mathbf{.004}$ & $\mathbf{.029}$ & $\mathbf{.001}$ & $\mathbf{.129}$
& $\mathbf{.000}$ & $\mathbf{.065}$ & $\mathbf{.000}$ & $\mathbf{.349}$
& $\mathbf{.003}$ & $\mathbf{.049}$ & $\mathbf{.000}$ & $\mathbf{.120}$ \\
\bottomrule
\end{tabular}
\end{adjustbox}
\caption{\textbf{ROC-AUC under adversarial attacks.} Persuasion attacks are
shown in \textbf{bold}; \ul{underlined} values indicate the strongest baseline
attack per column.}
\label{tab:auc_results}
\end{table*}

\clearpage
\section{Zero-Shot Verifier Robustness} \label{app:zero_shot}

A potential concern with fine-tuned verifiers is that the observed vulnerability could be an artefact of overfitting to the training distributions of FEVER and FEVEROUS. To rule this out, we evaluate a \textbf{zero-shot} verifier: the same \textsc{Llama-3-8B-Instruct} architecture used as one of our fine-tuned verifiers, but without any fine-tuning on FEVER or FEVEROUS, performing classification via prompting (see \autoref{fig:verifier_prompt}). By construction, this verifier cannot have overfitted to either training distribution. \autoref{tab:zero_shot} contrasts the fine-tuned \textsc{Llama-3-8B} from \autoref{tab:main_results} with its zero-shot counterpart.

\begin{table}[H]
\centering
\begin{adjustbox}{width=\columnwidth}
\begin{tabular}{llcccc}
\toprule
\textbf{Setting} & \textbf{Dataset} & \textbf{Verifier} & \textbf{Original} & \textbf{Blind} & \textbf{Oracle} \\
\midrule
Claim-Only & FEVER & Fine-tuned & $0.828$ & $0.623$ & $0.010$ \\
Claim-Only & FEVER & Zero-shot & $0.753$ & $0.599$ & $0.182$ \\
Claim-Only & FEVEROUS & Fine-tuned & $0.617$ & $0.518$ & $0.044$ \\
Claim-Only & FEVEROUS & Zero-shot & $0.549$ & $0.502$ & $0.167$ \\
\midrule
Gold-Evidence & FEVER & Fine-tuned & $0.903$ & $0.761$ & $0.127$ \\
Gold-Evidence & FEVER & Zero-shot & $0.893$ & $0.771$ & $\mathbf{0.153}$ \\
Gold-Evidence & FEVEROUS & Fine-tuned & $0.850$ & $0.765$ & $0.354$ \\
Gold-Evidence & FEVEROUS & Zero-shot & $0.782$ & $0.691$ & $\mathbf{0.146}$ \\
\bottomrule
\end{tabular}
\end{adjustbox}
\caption{Accuracy of fine-tuned vs.\ zero-shot \textsc{Llama-3-8B-Instruct} verifiers under persuasion attacks (Blind and Oracle).}
\label{tab:zero_shot}
\end{table}

The zero-shot verifier starts from a slightly lower clean baseline than its fine-tuned counterpart, as expected without task-specific adaptation, but its degradation profile under attack is comparable: under \textit{Blind}, Gold-Evidence accuracy drops to $0.771$ (zero-shot) vs.\ $0.761$ (fine-tuned) on FEVER, and to $0.691$ vs.\ $0.765$ on FEVEROUS; under \textit{Oracle}, both verifiers collapse below $0.20$ on FEVER and on FEVEROUS Claim-Only, with the zero-shot model in fact \textit{more} vulnerable than the fine-tuned one on FEVEROUS Gold-Evidence ($0.146$ vs.\ $0.354$). The persuasion-vs-baseline gap is also preserved: the \textit{Synonym Substitution}, \textit{Character Perturbations}, \textit{Word Swap}, and \textit{Paraphrasing} baselines leave zero-shot Gold-Evidence accuracy at $0.81$--$0.86$ on FEVER and $0.74$--$0.77$ on FEVEROUS. These results indicate that the observed vulnerabilities are not an artefact of fine-tuning on the benchmark training distributions, but instead reflect a more general sensitivity of LLM-based fact-checking systems to persuasion-based perturbations.

\section{Per-Technique Dense Retrieval Results} \label{app:dense_retrieval}

\autoref{tab:dense_per_technique} complements the aggregate retrieval results in \autoref{sec:results_rq3} with the per-technique retrieval degradation ($\Delta R@5 = R@5_{adv} - R@5_{orig}$) under blind persuasion, for both retrievers and both datasets, restricted to the $15$ validated techniques. The same techniques dominate the ranking under both retrieval paradigms: \textit{Obfuscation} and \textit{Slogan} are the most damaging across both retrievers and both datasets, with \textit{Appeal to Values}, \textit{Guilt by Association}, and \textit{Flag Waving} consistently in the next tier. Per-technique degradation is consistently less negative under Contriever, reflecting the dense retriever's partial robustness to blind persuasion discussed in \autoref{sec:results_rq3}.

\begin{table}[H]
\centering
\begin{adjustbox}{width=\columnwidth}
\begin{tabular}{lcccc}
\toprule
& \multicolumn{2}{c}{\textbf{FEVER}} & \multicolumn{2}{c}{\textbf{FEVEROUS}} \\
\cmidrule(lr){2-3}\cmidrule(lr){4-5}
\textbf{Technique} & \textbf{BM25} & \textbf{Contriever} & \textbf{BM25} & \textbf{Contriever} \\
\midrule
Obfuscation & $-0.528$ & $-0.420$ & $-0.555$ & $-0.451$ \\
Slogan & $-0.294$ & $-0.226$ & $-0.288$ & $-0.319$ \\
Appeal to Values & $-0.271$ & $-0.116$ & $-0.314$ & $-0.225$ \\
Guilt by Association & $-0.195$ & $-0.075$ & $-0.134$ & $-0.078$ \\
Flag Waving & $-0.164$ & $-0.061$ & $-0.119$ & $-0.071$ \\
Whataboutism & $-0.154$ & $-0.076$ & $-0.144$ & $-0.092$ \\
Loaded Language & $-0.144$ & $-0.024$ & $-0.039$ & $-0.025$ \\
Red Herring & $-0.142$ & $-0.027$ & $-0.085$ & $-0.026$ \\
Casting Doubt & $-0.124$ & $-0.018$ & $-0.102$ & $-0.070$ \\
Appeal to Hypocrisy & $-0.122$ & $-0.010$ & $-0.121$ & $-0.074$ \\
Conversation Killer & $-0.111$ & $-0.047$ & $-0.174$ & $-0.139$ \\
Name Calling / Labelling & $-0.101$ & $-0.068$ & $-0.073$ & $-0.096$ \\
Appeal to Popularity & $-0.090$ & $-0.011$ & $-0.129$ & $-0.081$ \\
Repetition & $-0.087$ & $-0.021$ & $-0.029$ & $-0.024$ \\
Appeal to Authority & $-0.075$ & $+0.006$ & $-0.072$ & $-0.048$ \\
\bottomrule
\end{tabular}
\end{adjustbox}
\caption{Per-technique retrieval degradation ($\Delta R@5$) under blind persuasion, sorted by FEVER BM25 (most damaging at the top). More negative values indicate more effective attacks.}
\label{tab:dense_per_technique}
\end{table}

\section{Examples of Evasion} \label{app:examples}

\autoref{tab:persistence_examples} presents representative examples of adversarial claims for the persuasion techniques identified as persistent threats in RQ2 (see \autoref{sec:results_rq2}). For each technique, we show one example from \textbf{FEVER} and one from \textbf{FEVEROUS}, illustrating how persuasive reframings can induce evasion even when the underlying factual content is contradicted by gold-evidence.

\begin{table*}[t]
\centering
\tiny
\renewcommand{\arraystretch}{1.3} 
\setlength{\tabcolsep}{4pt}       

\newcolumntype{L}{>{\raggedright\arraybackslash}X}
\newcolumntype{S}{>{\raggedright\arraybackslash}p{1.5cm}} 
\newcolumntype{M}{>{\raggedright\arraybackslash}p{1.8cm}} 

\begin{tabularx}{\textwidth}{S M L L L}
\toprule
\textbf{Persuasion Technique} & \textbf{Dataset} & \textbf{Original Claim} \newline \textit{(predicted False)} & \textbf{Adversarial Claim} \newline \textit{(predicted True)} & \textbf{Gold-Evidence} \\
\midrule

\multirow{9}{=}{\textbf{Obfuscation}} 
& FEVER 
& Hannah and Her Sisters' ending is \textbf{unrelated to Thanksgiving}.
& The conclusion of "Hannah and Her Sisters" has a nuanced \textbf{disconnect from the holiday season}.
& Hannah and Her Sisters is a 1986 American comedy-drama film which tells the intertwined stories of an extended family over two years that \ul{begins and ends with a family Thanksgiving dinner}.
 \\ \\
\addlinespace[0.5em] 
& FEVEROUS 
& The Malaysian plover lays eggs on beaches which are \textbf{incubated for 50 days}.
& The species incubates its eggs in coastal areas for a duration of \textbf{approximately one and a half months}.
& The eggs are incubated by both the male and female for \ul{about 30 days}, and then both parents care for the precocial chicks until they can fly after about 30 more days. \\
\midrule

\multirow{10}{=}{\textbf{Repetition}} 
& FEVER 
& The Hunger Games are series of \textbf{three films}. 
& The Hunger Games, The Hunger Games, The Hunger Games--are series of \textbf{three films}.
& The Hunger Games film series consists of \ul{four science fiction dystopian adventure films} based on The Hunger Games trilogy of novels, by the American author Suzanne Collins. \\
\addlinespace[0.5em]
& FEVEROUS 
& GZMB is a serine protease, an enzyme with \textbf{no responsibility for cleaving peptide bonds in proteins}.
& GZMB is a serine protease, a serine protease with \textbf{no responsibility for cleaving peptide bonds in proteins}.
& \textbf{1.} Granzyme B is a serine protease that in humans is encoded by the GZMB gene. \textbf{2.} Serine proteases (or serine endopeptidases) are enzymes that \ul{cleave peptide bonds in proteins}, in which serine serves as the nucleophilic amino acid at the (enzyme's) active site. \\
\midrule

\multirow{7}{=}{\textbf{Slogan}} 
& FEVER 
& The Nottingham Forest F.C. is \textbf{based in Lincolnshire, England}.
& "Nottingham Forest F.C.: \textbf{Rooted in Lincolnshire}, Shining in Every Corner!"
& Nottingham Forest Football Club is a professional association football club \ul{based in Nottingham, England}. \\
\addlinespace[0.5em]
& FEVEROUS 
& Steven Horsford \textbf{retired from the position} of Nevada Senate Majority Leader in 2009 and appointed a senator to lead the caucus election efforts during for the 2012 election cycle.
& "\textbf{Step aside} for a new era: Horsford \textbf{Hands the Baton} in 2009!"
& \ul{In February 2009, he assumed the position} of Nevada Senate Majority Leader. \\
\midrule

\multirow{8}{=}{\textbf{Flag Waving}} 
& FEVER 
& Honeymoon is the \textbf{third} major-label record by Lana Del Rey. 
& True fans of American music must recognize Honeymoon as the \textbf{third} major-label record by Lana Del Rey. & Honeymoon is the \ul{fourth studio album} and third major-label record by American singer and songwriter Lana Del Rey. \\
\addlinespace[0.5em]
& FEVEROUS  & The Galle Polling Division is in the Galle Electoral District in Sri Lanka, with a 2012 total population of 101,749 covering 11,860 square miles; and Galle matched the Parliamentary Election results of the country \textbf{five out of seven times}.
 & True Sri Lankans must support the accurate representation of their home in the Galle Polling Division, which matched the Parliamentary Election results \textbf{five out of seven times}.
 & The winner of Galle has matched the final country result \ul{5 out of 8 times}. \\
\midrule

\multirow{6}{=}{\textbf{Appeal to Popularity}} 
& FEVER
& One Dance was Drake's first number one single in \textbf{34 countries}. 
& Everyone is celebrating One Dance as Drake's first number one single in \textbf{34 countries}.
& ``One Dance'' reached number one in \textbf{15 countries}, including Australia... \\
\addlinespace[0.5em]
& FEVEROUS & The First Question Award (All tracks are written by Cornelius) released in February 25, 1994 is the debut studio album by Cornelius, which peaked at \textbf{number one} on the Oricon Albums Chart.
& The First Question Award, released in February 25, 1994, is the debut studio album by Cornelius—everyone is still celebrating its success at \textbf{number one} on the Oricon Albums Chart.
 & \textbf{1.} Japanese Albums (Oricon): 4.
  \textbf{2.} The First Question Award \ul{peaked at number four} on the Oricon Albums Chart. \\
\bottomrule
\end{tabularx}
\caption{Examples of effective persuasion techniques. Examples of adversarial claims for persuasion techniques identified as effective threats (see \autoref{sec:results_rq2}). Each example alters the claim to cause evasion: the evidence-based AFC model flips from a correct prediction (\textit{False}) in the original claim to an incorrect prediction (\textit{True}) in the adversarial claim. \textit{False} information is in \textbf{bold}, and direct counter evidence is \ul{underscored}.}
\label{tab:persistence_examples}
\end{table*}

Across techniques, the examples reveal a common pattern: persuasive modifications do not introduce new factual claims, but instead alter emphasis, framing, or salience in ways that reduce the model’s sensitivity to explicit contradictions in the evidence. \textit{Obfuscation} replaces concrete predicates with vaguer temporal or semantic expressions; \textit{Repetition} reiterates misleading elements to distract from the false predicate; and \textit{Slogan} and \textit{Flag Waving} inject rhetorical or identity-based language that shifts the claim away from a strictly factual register. \textit{Appeal to Popularity} further amplify this effect by introducing evaluative or social cues that the model appears to treat as non-factual context.

In all cases, the evidence remains sufficient to refute the claim, yet the evidence-based AFC model flips from a correct \textit{False} prediction on the original claim to an incorrect \textit{True} prediction on the adversarial variant.

\end{document}